\newcommand{\shortversion}[1]{}
\newcommand{\longversion}[1]{#1}

\documentclass[letterpaper, 10 pt, conference]{ieeeconf}  

\IEEEoverridecommandlockouts                              

\let\labelindent\relax

\usepackage{amsmath}
\usepackage{amsfonts}
\usepackage{algorithm} 
\usepackage{algorithmic}  
\usepackage[algo2e,linesnumbered,vlined,ruled]{algorithm2e}
\SetKwRepeat{Do}{do}{while}%
\usepackage{color}
\usepackage{lipsum}
\usepackage{tabularx}
\usepackage{xcolor}
\usepackage{multirow}
\usepackage{siunitx}
\usepackage{subcaption}
\usepackage{balance}
\usepackage{mathtools}
\usepackage{amsthm}
\usepackage{amssymb}
\usepackage[switch]{lineno}
\usepackage{siunitx}
\usepackage{ifthen}
\usepackage{xspace}
\usepackage{enumitem}
\usepackage{etoolbox}

\usepackage {tikz}
\usetikzlibrary{decorations.pathreplacing}
\usetikzlibrary{arrows,automata}
\usetikzlibrary {positioning}
\definecolor{lg}{RGB}{191,191,191}
\definecolor{g}{RGB}{144,144,144}
\definecolor{dg}{RGB}{47,79,79}
\definecolor{black}{RGB}{0,0,0}
\definecolor{myDarkRed}{RGB}{167,114,115}
\definecolor{myRed}{RGB}{255,58,70}
\definecolor{llb}{RGB}{135,206,235}
\definecolor{lb}{RGB}{0,191,255}
\definecolor{b}{RGB}{0,0,255}
\definecolor{mycolor}{rgb}{0.858, 0.188, 0.478}

\newcommand{\comment}[1]{}

\newcommand{\greedyone}{{\texttt{Greedy1}\ }}
\newcommand{\greedytwo}{{\texttt{Greedy2}\ }}

\theoremstyle{definition}
\newtheorem{theorem}{Theorem}
\newtheorem{lemma}{Lemma}
\newtheorem{problem}{Problem} 
\newtheorem{example}{Example}
\AtEndEnvironment{example}{\qed \vspace{-1mm}}

\makeatletter
\def\endthebibliography{%
  \def\@noitemerr{\@latex@warning{Empty `thebibliography' environment}}%
  \endlist
}
\makeatother

\shortversion{\linespread{0.905}}

\title{\LARGE \bf
\textsf{DT*}: Temporal Logic Path Planning in a Dynamic Environment
}

\author{Priya Purohit$^{1}$ and Indranil Saha$^{2}$
\noindent
\thanks{*The authors thankfully acknowledge the Defence Research Development Organisation (DRDO), India for funding the project through JCBCAT, Kolkata}
\thanks{$^{1}$ Priya Purohit is with Department of Computer Science and Engineering,
Indian Institute of Technology Kanpur
{\tt\small priyapr@cse.iitk.ac.in}}%
\thanks{$^{2}$ Indranil Saha is with Department of Computer Science and Engineering,
Indian Institute of Technology Kanpur
{\tt\small isaha@cse.iitk.ac.in}}%
}

\begin{document}

\maketitle
\thispagestyle{empty}
\pagestyle{empty}

\begin{abstract}
Path planning for a robot is one of the major problems in the area of robotics. When a robot is given a task in the form of a Linear Temporal Logic  (LTL) specification such that the task needs to be carried out repetitively, we want the robot to follow the shortest cyclic path so that the number of times the robot completes the mission within a given duration gets maximized. 
In this paper, we address the LTL path planning problem in a dynamic environment where the newly arrived dynamic obstacles may invalidate some of the available paths at any arbitrary point in time. We present $\textsf{DT*}$, an SMT-based receding horizon planning strategy that solves an optimization problem repetitively based on the current status of the workspace to lead the robot to follow the best available path in the current situation. We implement our algorithm using the Z3 SMT solver and evaluate it extensively on an LTL specification capturing a pick-and-drop application in a warehouse environment. We compare our SMT-based algorithm with two carefully crafted greedy algorithms. Our experimental results show that the proposed algorithm can deal with the dynamism in the workspace in LTL path planning effectively.
\end{abstract}


\section{Introduction}
\label{sec-intro}

Using Linear Temporal Logic (LTL)~\cite{BaierK08} as a formal specification language is a convenient way to capture complex requirements for a mobile robot. Linear temporal logic enables one to capture those requirements that entail that the robot remains operational for a long time to carry out repetitive work.
Several techniques can be employed to synthesize an infinite-length trajectory from a given LTL specification~\cite{KaramanF09,BhatiaKV10,PatriziLGG11,UlusoySDBR13,WolffTM14}.
An infinite-length trajectory satisfying an LTL formula can be represented as a prefix followed by a loop that can be unrolled to generate a perpetual behavior. Thus, while following a trajectory satisfying an LTL formula, the robot needs to reach a loop and follow it repetitively.

For a complex robotic system, a robot may have the option to choose one of the multiple possible loops that allow the robot to satisfy the requirement. 
The efficiency of the robot depends on how quickly it can cover a loop as the throughput of the robot is measured by the number of times it completes the loop within a given duration. 
For example, consider an application of warehouse management where a robot is employed to perform a pick-and-drop operation~\cite{smith2}.
Suppose that the robot has to pick an object from one of the three racks that are located at three different locations in the workspace and bring it to one of the two different drop locations. 
Thus, the robot has the option of following six loops to satisfy the requirement. The robot follows the shortest loop to maximize its efficiency.

The situation becomes challenging when some of the loops may not be available due to some dynamic events in the workspace. 
In our warehouse example, some racks may not be available to the robot due to some ongoing service by some human operator. 
If the loop that the robot was traversing becomes inaccessible, the robot may choose to wait to get access to it eventually. 
Alternatively, it may switch to another suitable loop and keep satisfying the LTL specification. 
The robot may have information about the dynamic events in the workspace, which may help it take the decision to switch to an appropriate loop. 
For example, in the use-case of warehouse management, the human operators may communicate with the robot the time instance when she will start servicing a rack and the approximate duration she would take to complete this operation. 
However, in the presence of many possible loops and several dynamic events happening in the workspace, it is algorithmically challenging for the robot to decide the optimal course of action at any given point in time.


In this paper, we propose $\textsf{DT*}$, a solution to the above-mentioned problem through a reduction to SMT (Satisfiability Modulo Theory) solving problems~\cite{BarST-SMTLIB}. In this  approach, we use the \emph{reduced product graph} introduced as part of the $\textsf{T*}$ algorithm~\cite{KhalidiSaha18} to encode the trajectory of the robot by treating the loops to be taken at different steps as the decision variables. 
Taking inspiration from 
receding horizon motion planning in dynamic environments~\cite{UlusoyB14,WongpiromsarnTM12}, we employ a receding horizon mechanism where the planning is carried out for a horizon starting from the current time point. 

We perform an extensive simulation to evaluate $\textsf{DT*}$. Through a comparison with the two greedy algorithms, we demonstrate that despite the computational overhead, $\textsf{DT*}$ can enable a robot to achieve much superior performance in a dynamic environment.
We demonstrate the practical applicability of our algorithm to a real robotic system through a  simulation on ROS~\cite{ROS}.

In summary, we make the following contributions: 
\begin{itemize}
    \item We introduce the online LTL motion planning problem in a dynamic environment and propose $\textsf{DT*}$ an SMT-based algorithm to solve the problem.
    \item We evaluate $\textsf{DT*}$ extensively through a comparison with two greedy algorithms.
    \item We provide a ROS-based simulation to demonstrate how our algorithm will be operational to solve the online LTL motion planning problem in a dynamic environment in practice.
\end{itemize}

\section{PROBLEM}
\subsection{Preliminaries}

\subsubsection{Workspace and Actions.}
We represent the workspace $W$ as a 2D rectangular grid environment. Each cell of the grid is referenced by its $x,y$ coordinates. Some of the cells in the workspace may be occupied by obstacles. The motion of the robot within the workspace is captured by a set of actions $Act$. 
For a 2D workspace, $Act$ could be $\mathtt{left}$, $\mathtt{right}$, $\mathtt{up}$, $\mathtt{down}$. The cost associated with an action denotes the time taken to execute the action.
Though we present our framework and the experiments on 2D environments, our framework can be extended seamlessly to 3D environments.

\subsubsection{Weighted Transition System}
Let $T$ be the transition system modeling the motion of a robot in $W$, which is defined as $T:=\ (S_T, O_T, S_{T_0}, E, \Pi, L_T, w_T)$. Here,
(i) $S_T$ denotes the set of all cells in $W$,
(ii) $O_T \subset S_T$ is the set of cells in $W$ that are occupied by obstacles, 
(iii) $S_{T_0} \in S_T \setminus O_T$ is the initial state,
(iv) $E \subseteq$ $(S_T \setminus O_T) \times (S_T \setminus O_T)$ is the state transitions, for $s_1, s_2 \in (S_T \setminus O_T)$, $(s_1,s_2) \in E$ iff there  exist an $act \in Act$ such that $s_{1} \stackrel{act}{\longrightarrow} s_{2}$.
(v) $\Pi$ denotes the set of all atomic propositions,
(vi) $L_T$ : $S\ \rightarrow 2^\Pi$ maps the states in $S$ to the propositions $\mathtt{true}$ at that state, and
(vii) $w_T$ : $E \rightarrow \mathbb{R}_{>0}$ is a function capturing the cost of the action on an edge $e\in E$. 


\subsubsection{Linear Temporal Logic}
Temporal logic extends Propositional logic by capturing the notion of time \cite{book2}. Linear Temporal Logic (LTL) contains all the standard Boolean operators in the propositional logic (i.e., $\mathtt{T}$ ($\mathtt{true}$), $\neg$ (negation), and $\wedge$ (conjunction)). Along with these operators, LTL also contains temporal operators $\bigcirc$ (Next) and $\mathtt{U}$ (Until)~\cite{book1}. 
The Next operator $\bigcirc$ is a unary operator and is followed by a formula, which is observed in the next time-step. 
The Until operator $\mathtt{U}$ is a binary operator between two formulas. The formula $\phi_1  \mathtt{U}\, \phi_2$ says that $\phi_2$ should be observed at some step $k$, and for all steps $t$, $0 \leq t <k$, $\phi_1$ must be observed.
There are two other widely-used temporal operators, namely $\Diamond$ (Eventually) and $\square$ (Always), which can be derived from the basic logical and temporal operators as follows:
$\Diamond \phi:=\mathtt{T}\, \mathtt{U}\, \phi$ and
$\square \phi:=\neg \Diamond \neg \phi$.
Here, $\Diamond \phi$ says that $\phi$ will be observed at some time-step eventually, and  $\square \phi$ says that $\phi$ will be observed at all the steps, i.e., it is not the case that $\neg\phi$ will be observed eventually.

\subsubsection{B\"uchi Automaton}
Given an LTL specification $\phi$, a B\"uchi Automaton $B_{\phi}$ models $\phi$. 
A B\"uchi automaton is represented as a tuple $B_{\phi}=\left(S_B, S_{B_0}, O, \delta, F\right),$ where
(i) $S_B$ is a finite set of states, 
(ii) $S_{B_0} \subseteq S_B$ is the set of initial states,
(iii) $O$ is the set of input alphabets,
(iv) $\delta: S_B \times O \rightarrow S_B$ is a transition function, and
(v) $F \subseteq S_B$ is the set of accepting (final) states.
A run over an infinite input sequence $w(o)$ = $s_0\, s_1 \ldots$ is a sequence of automata states $\rho$ = $ q_0\, q_1 \ldots$, with $q_0 \in S_{B_0}$ and $q_0\xrightarrow[]{s_0}q_1,\ q_1\xrightarrow[]{s_1}q_2$ and so on, where $s_i \in O$.
An infinite input sequence $w(o)$ is said to be accepted by B\"uchi Automaton $B$ iff there exists at least one run in which at least one state in $F$ is visited infinitely often.

\subsubsection{Product Graph}
The product graph $P$ of the transition system $T$ and B\"uchi automaton
$B_{\phi}$ is defined as: $P =\left(S_{P}, S_{P_0}, E_{P}, F_{P}, w_{p}\right)$,
where, (i) $S_P=S_T \times S_B$, 
(ii) $S_{P_0}$ = $S_{T_0} \times S_{B_0}$, 
(iii) $E_P \in S_P \times S_P$ , where $((s_i,q_i),(s_j,q_j))$ $\in E_P$ iff $(s_i,s_j)\in E$ and $\exists\ c \in 2^{\Pi}$, $\delta (q_i,c) = q_j$ such that $c\in L_T(s_j)$,
(iv) $F_P =  S_T \times F$, and
(v) $w_p$ : $E_P \rightarrow \mathbb{R}_{>0}$, such that $w_P((s_i,q_i),(s_j,q_j))$ = $w_T(s_i,s_j)$.

\subsubsection{Reduced Product Graph}
A reduced graph $G_r$ of a product graph $P$ is defined by:
\mbox{$G_r$ = $(V_r,v_0,E_r,F_r,w_r)$}, where
(i) $V_r \subseteq S_T \times S_B$,
(ii) \mbox{$v_0 = S_{T_0} \times S_{B_0}$},
(iii) $E_r \subseteq V_r \times V_r$, 
(iv) $F_r \subseteq V_r$, $(s_i,q_i) \in F_r$ iff $q_i \in F$, and 
(v) $w_{r}: E_{r} \rightarrow \mathbb{R}_{>0},$ a weight function.

The reduced graph differs from the original product graph as it may add direct edges between $(s_i,q_i)$ to $(s_j,q_j)$ even when $(s_i,s_j) \notin E$. 
An edge in $G_r$ may represent a path in the product graph $P$.
To compute the path length between any adjacent edge of $G_r$, we can use the A* algorithm~\cite{astar}.
For more information, readers are encouraged to refer~\cite{Tstar}.



\begin{figure}
\centering
\captionsetup[subfigure]{justification=centering}
\tabskip=0pt
\valign{#\cr
\hbox{%
\begin{subfigure}[b]{.10\textwidth}
\centering

\resizebox{5cm}{5cm}{
\begin{tikzpicture}[->,>=stealth',shorten >=1pt,auto,semithick,node distance=3cm, edge_style/.style={draw=black, semithick} ]

  \tikzset{%
    in place/.style={  auto=false, fill=white, inner sep=2pt, },}
     \node[state,initial] (q0) {\Large $q_0$};
     \node[state, above right  of=q0] (q1) {\Large $q_1$};
     \node[state, below right  of=q0] (q2) {\Large $q_2$};
     \node[state,  below right of=q1] (q3) {\Large $q_3$};
     \node[state,accepting,  below right of=q1] () {};
       \draw[->, auto]
    (q0) edge[loop above] node {\small $!p1\&!p2$} (q0)
    (q0) edge node[pos=0.8] {\small $p1\&!p2$} (q1)
    (q0) edge node[pos=0.8,swap] {\small $!p1\&p2$} (q2)
    (q2) edge[loop below] node[in place, swap] {\small $!p1\&!p2$} (q2)
    (q2) edge node[pos=0.6, in place, swap] {\small $p1\&!p2$} (q1)
    (q1) edge[loop above] node[in place] {\small $!p1\&!p2$} (q1)
    (q1) edge[bend right=20] node[pos=0.6, in place, swap] {\small $!p1\&p2$} (q3)
    (q3) edge node {\small $!p1\&!p2$} (q2)
    (q3) edge[bend right=20] node[pos=0.7, in place, swap] {\small $p1\&!p2$} (q1)
    ;
\end{tikzpicture}}

    \caption{}
    \label{fig1.1}
    \end{subfigure}%
  } \cr
  \noalign{\hfill}
  \hbox{%
    \begin{subfigure}{.2\textwidth}
    \centering
    \includegraphics[height=2.2cm, width=2.5cm]{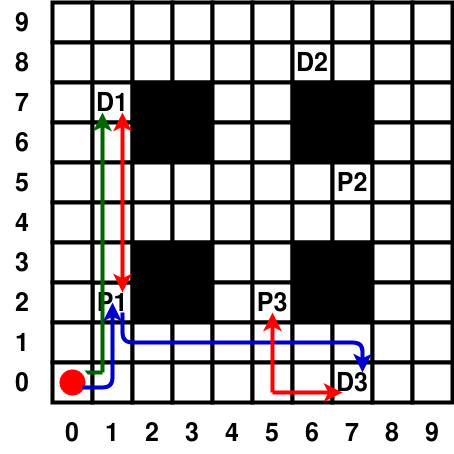}
    \caption{}
    \label{fig1.2}
    \end{subfigure}%
  }\vfill
  \hbox{%
    \begin{subfigure}{.2\textwidth}
    \centering
    \includegraphics[height=2.2cm, width = 2.5cm]{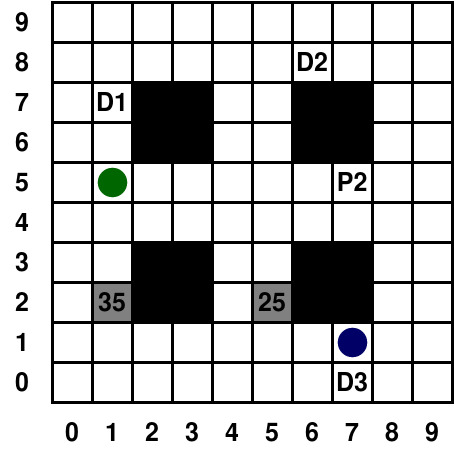}
    \caption{\ }
    \label{fig1.3}
    \end{subfigure}%
  }\cr
}
\caption{(a) B\"uchi Automaton for LTL specification: $\Box(\Diamond p1 \wedge \Diamond p2) \wedge \Box ((p1 \rightarrow \bigcirc (\neg p1 \, \mathtt{U}\, p2)) \wedge  (p2 \rightarrow \bigcirc(\neg p2\, \mathtt{U}\, p1))$,   
(b) the plans generated by Greedy algorithms,
(c) the environment changes occurring at timestamp 10.}
\label{fig1}
\end{figure}
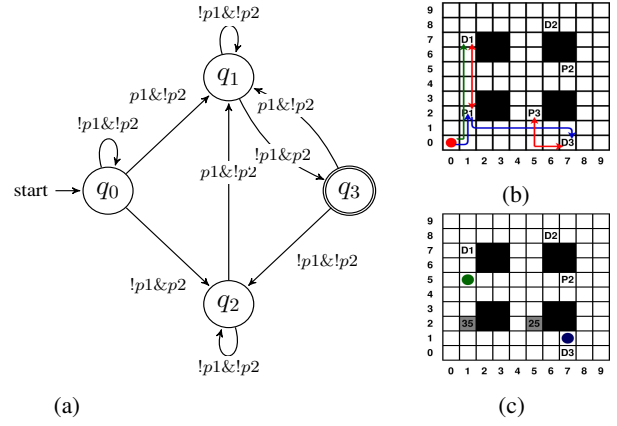

\subsubsection{Robot Trajectory}
Consider a robot whose motion in a workspace $W$ is modeled by a transition system $T$.
Suppose that the robot is given a task that needs to be repeated in the form of an LTL query $\phi$, which is represented by a B\"uchi automaton $B_{\phi}$. We create $P$, the product graph between the transition system $T$ modeling workspace $W$ and $B_{\phi}$ modeling the task specification. 
The nodes of this graph can be represented by $(x,y,s)$, where $(x,y)\in S_T$ and $s \in S_B$. 
Our task to satisfy $\phi$ in $W$ can then be modeled as finding a cycle consisting of any final state $f\in F_P$ in the product graph $P$. 
As one can see, multiple such cycles could be possible in the product graph. 

We call the path from the initial robot location to one of the final locations of the graph as the \emph{prefix path} $R_{pref}$ and from the final location to itself as the \emph{suffix cycle} $R_{suff}$.
The infinite run $R$ over $T$ can then be written as $R$ = $R_{pref}$.$(R_{suff})^\omega$, where $R_{pref}$ is traversed once and $R_{suff}$ is traversed infinitely. As the LTL specification requires the task  to be repeated periodically, an optimal solution for an infinite run $R$ has the suffix cycle with the minimum cost.

\subsection{Problem Statement and Naive Solutions}
To satisfy an LTL specification $\phi$, the suffix cycle $R_{suff}$ has to pass through some locations where some atomic propositions hold $\mathtt{true}$.
Consider a situation where the shortest suffix cycle is no longer available due to some dynamic obstacle. The dynamic obstacle may capture the position corresponding to an atomic proposition on the suffix cycle by making it unavailable. 
It may also be present at some other location (not corresponding to a proposition) on the suffix cycle and make it completely unavailable or increase its length to the extent that it is no longer the shortest suffix cycle.
We assume that the duration for which a dynamic obstacle will keep a suffix cycle unavailable is known to the robot.
In the above context, we define the following problem.

\begin{problem}
Given a robot transition system $T$ and an LTL specification $\phi$, design an online algorithm that in the presence of dynamic changes in the environment generates the trajectory of the robot satisfying the following:
\begin{itemize}
    \item There exists an infinite extension of the current trajectory that satisfies the LTL formula $\phi$.
    \item At any time point $T$, the number of loops covered by the trajectory in the duration $[0, T]$ gets maximized.
\end{itemize}
\end{problem}

\begin{figure}
\begin{subfigure}{{\textwidth}}
\begin{tikzpicture}[%
    every node/.style={
        font=\scriptsize,
        text height=1ex,
        text depth=.25ex,
    },
]
\draw[->] (0,0) -- (7.5,0);
\node[anchor=north] at (0,0) {$10$};
\node[anchor=north] at (1.5,0) {$20$};
\node[anchor=north] at (2.3,0) {$25$};
\node[anchor=north] at (3.4,0) {$33$};
\node[anchor=north] at (4.2,0) {$37$};
\node[anchor=north] at (5,0) {$41$};
\node[anchor=north] at (5.8,0) {$45$};
\node[anchor=north] at (6.6,0) {$49$};
\node[anchor=south] at (0,0) {$(7,1)$};
\node[anchor=south] at (1.5,0) {$(6,8)$};
\node[anchor=south] at (2.3,0) {$(8,5)$};
\node[anchor=south] at (3.4,0) {$(7,0)$};
\node[anchor=south] at (4.2,0) {$(5,2)$};
\node[anchor=south] at (5,0) {$(7,0)$};
\node[anchor=south] at (5.8,0) {$(5,2)$};
\node[anchor=south] at (6.6,0) {$(7,0)$};
\node[anchor=south] at (7.4,0) {Loc};
\node[anchor=north] at (7.4,0) {Time};

\draw[dg,dashed,thick,-latex] (0,-0.5) -- (1.5,-0.5);
\draw[dg,dashed,thick,-latex] (1.5,-0.5) -- (3.4,-0.5);

\fill[dg] (3.4,-0.4) rectangle (4.2,-0.55);
\fill[g] (4.2,-0.4) rectangle (5,-0.55);
\fill[dg] (5,-0.4) rectangle (5.8,-0.55);
\fill[g] (5.8,-0.4) rectangle (6.6,-0.55);

\draw[decorate,decoration={brace,amplitude=5pt}] (0,0.45) -- (1.5,0.45)
    node[anchor=south,midway,above=4pt] {Prefix Path1};
\draw[decorate,decoration={brace,amplitude=5pt}] (2.3,0.45) -- (3.4,0.45)
    node[anchor=south,midway,above=4pt] {Prefix Path2};
\draw[decorate,decoration={brace,amplitude=5pt}] (3.5,0.45) -- (5,0.45)
    node[anchor=south,midway,above=4pt] {Suffix Cycle 1};
\draw[decorate,decoration={brace,amplitude=5pt}] (5,0.45) -- (6.6,0.45)
    node[anchor=south,midway,above=4pt] {Suffix Cycle 2};    
\end{tikzpicture}
\caption{\ \ \ \ \ \ \ \ \ \ \ \ \ \ \ \ \ \ \ \ \ \ \ \ \ \ \ \ \ \ \ \ \ \ \ \ \ \ \ \ \ \ \ \ \ \ \ \ \ \ \ \ \ \ \ \ \ \ \ \ \ \ \ \ \ \ \ \ \ \ \ \ \ \ \ \ \ \ \ \ \ \ \ \ }
\end{subfigure}
\begin{subfigure}{\textwidth}
\begin{tikzpicture}[%
    every node/.style={
        font=\scriptsize,
        text height=1ex,
        text depth=.25ex,
    },
]
\draw[->] (0,0) -- (7.5,0);

\node[anchor=north] at (0,0) {$10$};
\node[anchor=north] at (2.4,0) {$23$};
\node[anchor=north] at (3.1,0) {$27$};
\node[anchor=north] at (3.8,0) {$31$};
\node[anchor=north] at (4.5,0) {$35$};
\node[anchor=north] at (5.2,0) {$39$};
\node[anchor=north] at (5.9,0) {$43$};
\node[anchor=north] at (6.6,0) {$47$};
\node[anchor=south] at (0,0) {$(1,5)$};
\node[anchor=south] at (2.4,0) {$(7,0)$};
\node[anchor=south] at (3.1,0) {$(5,2)$};
\node[anchor=south] at (3.8,0) {$(7,0)$};
\node[anchor=south] at (4.5,0) {$(5,2)$};
\node[anchor=south] at (5.2,0) {$(7,0)$};
\node[anchor=south] at (5.9,0) {$(5,2)$};
\node[anchor=south] at (6.6,0) {$(7,0)$};
\node[anchor=south] at (7.4,0) {Loc};
\node[anchor=north] at (7.4,0) {Time};

\draw[dg,dashed,thick,-latex] (0,-0.5) -- (2.4,-0.5);


\fill[dg] (2.4,-0.4) rectangle (3.1,-0.55);
\fill[g] (3.1,-0.4) rectangle (3.8,-0.55);
\fill[dg] (3.8,-0.4) rectangle (4.5,-0.55);
\fill[g] (4.5,-0.4) rectangle (5.2,-0.55);
\fill[dg] (5.2,-0.4) rectangle (5.9,-0.55);
\fill[g] (5.9,-0.4) rectangle (6.6,-0.55);
\draw[decorate,decoration={brace,amplitude=5pt}] (0,0.45) -- (2.4,0.45)
    node[anchor=south,midway,above=4pt] {Prefix Path};
\draw[decorate,decoration={brace,amplitude=5pt}] (2.4,0.45) -- (3.8,0.45)
    node[anchor=south,midway,above=4pt] {SC1};
\draw[decorate,decoration={brace,amplitude=5pt}] (3.8,0.45) -- (5.2,0.45)
    node[anchor=south,midway,above=4pt] {SC2}; 
    \draw[decorate,decoration={brace,amplitude=5pt}] (5.2,0.45) -- (6.6,0.45)
    node[anchor=south,midway,above=4pt] {SC3};
    
\end{tikzpicture}
\caption{\ \ \ \ \ \ \ \ \ \ \ \ \ \ \ \ \ \ \ \ \ \ \ \ \ \ \ \ \ \ \ \ \ \ \ \ \ \ \ \ \ \ \ \ \ \ \ \ \ \ \ \ \ \ \ \ \ \ \ \ \ \ \ \ \ \ \ \ \ \ \ \ \ \ \ \ \ \ \ \ \ \ \ \ } 
\end{subfigure}
\begin{subfigure}{\textwidth}
\begin{tikzpicture}[%
    every node/.style={
        font=\scriptsize,
        text height=1ex,
        text depth=.25ex,
    },
]
\draw[->] (0,0) -- (7.5,0);
\node[anchor=north] at (0,0) {$10$};
\node[anchor=north] at (0.7,0) {$12$};
\node[anchor=north] at (2.3,0) {$19$};
\node[anchor=north] at (4,0) {$26$};
\node[anchor=north] at (5,0) {$34$};
\node[anchor=north] at (6,0) {$42$};
\node[anchor=north] at (7,0) {$50$};
\node[anchor=south] at (0,0) {$(7,1)$};
\node[anchor=south] at (0.7,0) {$(7,0)$};
\node[anchor=south] at (2.3,0) {$(7,5)$};
\node[anchor=south] at (4,0) {$(7,0)$};
\node[anchor=south] at (5,0) {$(7,0)$};
\node[anchor=south] at (6,0) {$(7,0)$};
\node[anchor=south] at (7.0,0) {$(7,0)$};
\node[anchor=south] at (7.6,0) {Loc};
\node[anchor=north] at (7.6,0) {Time};
\draw[dg,dashed,thick,-latex] (0,-0.5) -- (0.5,-0.5);

\fill[dg] (0.5,-0.4) rectangle (2.3,-0.55);
\fill[g] (2.3,-0.4) rectangle (4,-0.55);
\fill[dg] (4,-0.4) rectangle (4.5,-0.55);
\fill[g] (4.5,-0.4) rectangle (5,-0.55);
\fill[dg] (5,-0.4) rectangle (5.5,-0.55);
\fill[g] (5.5,-0.4) rectangle (6,-0.55);
\fill[dg] (6,-0.4) rectangle (6.5,-0.55);
\fill[g] (6.5,-0.4) rectangle (7.0,-0.55);

\draw[decorate,decoration={brace,amplitude=5pt}] (0,0.45) -- (0.5,0.45)
    node[anchor=south,midway,above=4pt] {Prefix Path};
\draw[decorate,decoration={brace,amplitude=5pt}] (0.5,0.45) -- (4,0.45)
    node[anchor=south,midway,above=4pt] {Suffix Cycle 1};
\draw[decorate,decoration={brace,amplitude=5pt}] (4,0.45) -- (5,0.45)
    node[anchor=south,midway,above=4pt] {SC2}; 
    \draw[decorate,decoration={brace,amplitude=5pt}] (5,0.45) -- (6,0.45)
    node[anchor=south,midway,above=4pt] {SC3};
    \draw[decorate,decoration={brace,amplitude=5pt}] (6,0.45) -- (7.0,0.45)
    node[anchor=south,midway,above=4pt] {SC4};
\end{tikzpicture}
\caption{\ \ \ \ \ \ \ \ \ \ \ \ \ \ \ \ \ \ \ \ \ \ \ \ \ \ \ \ \ \ \ \ \ \ \ \ \ \ \ \ \ \ \ \ \ \ \ \ \ \ \ \ \ \ \ \ \ \ \ \ \ \ \ \ \ \ \ \ \ \ \ \ \ \ \ \ \ \ \ \ \ \ \ \ }
\end{subfigure}
\caption{Timelines for (a) the plan generated by \greedyone Algorithm, (b) the plan generated by \greedytwo Algorithm, (c) an optimal plan for the environment given in Figure~\ref{fig1}}
\label{fig2}
\end{figure}
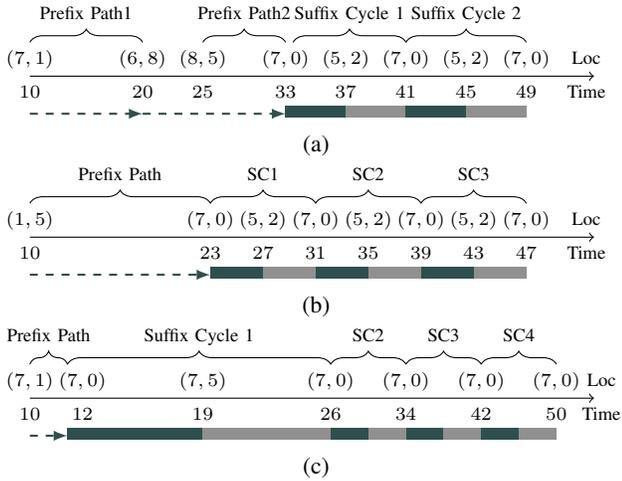

Here we mention two straightforward greedy solutions to the above-mentioned problem. Throughout the paper, we refer to them as \greedyone and \greedytwo.

\smallskip
\noindent
\textbf{\texttt{Greedy1}.} This greedy solution finds the shortest suffix cycle in the modified product graph in case the current suffix $R_{suff}$ gets invalidated due to a dynamic event in the environment and gets an infinite run $R$ of the form $R'_{pref}$. ${(R'_{suff})}^\omega$.
This algorithm gives us a plan where the length of the suffix cycle is minimized. 
However, this solution does not consider the fact that the proposition location may become available in some time, and it could be better to wait for proposition locations to become available or to switch to some nearby cycle from which it incurs less cost to return to the shortest cycle later.
 
\smallskip
\noindent
 \textbf{\texttt{Greedy2}.} This strategy minimizes the time within which one cycle can be completed to the earliest based on the current workspace. It minimizes the overall completion time of one prefix and the corresponding suffix and may choose a nearby longer cycle with a shorter prefix length.
 

\subsection{Example}
\label{sec-example}
To understand the complexity of the problem and the limitations of the above-mentioned greedy algorithms, let us consider an example.
Figure~\ref{fig1.1} shows the B\"uchi automaton for an LTL specification that captures a warehouse scenario where the robot is given a task that needs to be repeated forever: \textit{The robot should pick up an object from some location and drop it to some other designated location. Once an object is picked up, the robot cannot go to a pickup location again until it visits some drop location, and once an object is dropped, it cannot go to a drop location until it visits some pickup location.}
In figure~\ref{fig1.1}, the propositions $\mathtt{pickup}$ and $\mathtt{drop}$ are marked as $p1$ and $p2$ respectively.
Figure~\ref{fig1.2} shows a sample workspace with 3 pickup and 3 drop locations marked as $\mathtt{P1}$, $\mathtt{P2}$, $\mathtt{P3}$ and $\mathtt{D1}$, $\mathtt{D2}$, $\mathtt{D3}$, respectively. 
Figure~\ref{fig1.2} shows the initial solutions generated by \greedyone marked in blue (prefix) and red (suffix) and \greedytwo algorithm marked in green (prefix) and red (suffix).

Let us assume that at timestamp (henceforth, written as $ts$) $10$, proposition locations $\mathtt{P3}$ and $\mathtt{P1}$ becomes unavailable till $ts=25$ and $ts=35$ respectively.
The current robot position for both the paths generated by \greedyone and \greedytwo algorithms are shown using blue and green dots, respectively.

At $ts=10$, \greedyone algorithm provides cycle $\langle (6,8),(7,5),(6,8)\rangle$ as the minimum length cycle. 
But at $ts = 25$, $\mathtt{P3}$ becomes available again, \greedyone then replans at $ts=25$. As the robot in this case has already dropped the object at $(6,8)$, it first goes to pickup location $(7,5)$ again to obey the specification captured by $\phi$ before moving to the drop location at $(7,0)$. 
\greedytwo algorithm minimizes the overall time to complete a prefix followed by the corresponding suffix. The minimum time required to complete any cycle could be achieved by switching to cycle $\langle (7,0),(5,2),(7,0)\rangle$. The paths generated by \greedyone and \greedytwo algorithm are shown as timelines in Figure~\ref{fig2}(a) and ~\ref{fig2}(b) respectively. 

However, there exists a better plan which could complete even more cycles within the same duration.
Consider the robot location to be blue dot (same as \greedyone algorithm) at $Ts=10$. As shown in figure~\ref{fig2}(c), switching to a longer nearby cycle followed by completing shorter cycles could result in the completion of $4$ cycles till $ts=50$.
As shown in figure~\ref{fig2}, the number of times the optimizing task is completed till $ts=50$ are $2$, $3$ and $4$ with the above plans.

Through this example, we demonstrated how different algorithms would result in different decision sequences. The decision to be made by the robot is to choose whether to switch to another cycle, to repair the current trajectory, or simply wait for the proposition locations to become available again.
In this paper, we attempt to solve this problem by reducing it to a series of optimization problems, as described in the following section.

\section{$\textsf{DT*}$ Algorithm}

When we deal with the planning problem in an uncertain environment, the notion of a static plan is ruled out because the plan may get invalidated once some change in the environment is observed. 
Thus, we adopt a receding horizon planning strategy (motivated by~\cite{WongpiromsarnTM12}) in $\textsf{DT*}$, where the plan is computed within some finite horizon denoted by $H$.
\begin{algorithm2e}[!t]
\SetKwInOut{Input}{input}
\SetKwInOut{Output}{output}
\begin{small}
\DontPrintSemicolon
\Input{A transition system $T$ and LTL query $\phi$}
\Output{Dynamic plans based on the environment changes} 

$B_\phi \gets\mathtt{LTL2BA}(\phi)$\; \label{ltl2ba}
$G_r (V_r,v_0,E_r,F_r,w_r)\gets$ $\mathtt{gen\_redc\_graph}$($B_\phi$,$T$)\;\label{get_gr}
$pos_{cur} \gets v_0$, 
$time_{cur} \gets 0$, 
$time_{comp} \gets \nu$, 
$R \gets \emptyset$\;
$R\gets\mathtt{static\_plan}(G_r,T)$\; \label{static_plan}
\While{$True$}
{ 
  \While{change is not observed and $!R.empty()$}
  {  
    $pos_{next} \gets R.\mathtt{pop}()$\;\label{get_pos}
    $\mathtt{move\_robot}(pos_{cur}, pos_{next})$\;\label{move}
    $pos_{cur} \gets pos_{next}$\;\label{next_pos}
}
$D\gets \mathtt{env\_changes}(W)$\;\label{env_changes}
$G_r\gets\mathtt{update\_graph}(G_r, pos_{cur})$\;\label{atg}
$R \gets \emptyset$\;
\While{$R.empty()$}
{  
    $time_{cur} \gets time_{cur}+time_{comp}$\; \label{update_time}
    $Ed_{cost}\gets\mathtt{dy\_cost}$($G_r, pos_{cur}$, $time_{cur},H,T,D$)\;\label{dy_cost}
    $R\gets\mathtt{plan\_in\_H}(G_r, pos_{cur}, time_{cur}, H, Ed_{cost})$\;\label{plan_in_H}
}
}
%



\BlankLine
\BlankLine
\SetKwProg{myproc}{Procedure}{}{}\label{p2}
\myproc{ \textbf{plan\_in\_H}$(G_r, pos_{cur}, time_{cur}, H, Ed_{cost})$ } {
$cons \gets  
\mathtt{gen\_cons} (G_r, pos_{cur}, time_{cur}, H, Ed_{cost})$\;\label{get_cons}
$model \gets \mathtt{solve\_constraints}(cons)$\;\label{solve_cons}
\If{($model \ne \emptyset$)}
{ 
$R \gets \mathtt{get\_plan\_from\_model}(model)$\;\label{get_plan}
\Return $R$\; 
}
\Return $\emptyset$
}
\caption{$\textsf{DT*}$} 
\label{alg}
\end{small}
\end{algorithm2e}

Algorithm 1 outlines the major steps in $\textsf{DT*}$. Given a workspace $W$ modeled by $T$ and an LTL query $\phi$, we want to generate dynamic plans for the robot so that the robot can deal with the changes in the environment.
As the size of the transition system grows or the LTL query becomes more complex, the size of the product graph increases, and so does the time for path computation. To handle this scalability issue, we use a reduced graph $G_r$ of the original product graph $P$. 
For the generation of the reduced graph $G_r$, we use a procedure given in~\cite{Tstar} denoted by $\mathtt{gen\_redc\_graph}$ on line~\ref{get_gr}. 

In the initial static environment, we generate a plan of the form $R = R_{pref}.(R_{suff})^\omega$ on line~\ref{static_plan}, where $R_{pref}$ is the path from the initial location to a destination location $d \in F_r$, such that the cycle $R_{suff}$ from $d$ is the minimum length cycle. This suffix is repeated perpetually if the environment does not change or the horizon does not come to the completion (line~\ref{get_pos}-\ref{next_pos}).

Once one of these two events happens, we mark the environment changes in $W$ on line~\ref{env_changes}, if any. These environment changes capture the duration of unavailability of some grid locations of $W$. The cost of some of the actions $act\in Act$ of $T$ increases for these grid locations, as this cost now has to incorporate the waiting time. 
On line~\ref{atg}, we update $G_r$ based on the current robot state $(x_i,y_i,s_i)$.

For this updated graph $G_r$, we generate each edge's cost within horizon $H$ starting from $time_{cur}$. Method $dy\_cost$ on line~\ref{dy_cost} takes $G_r$, $T$, the environment changes captured by $D$ etc. as input. This method pre-computes all the edge costs of $G_r$ within horizon $H$. This method is an adaptation of the technique presented in \cite{llastar} to deal with dynamic environments. For path calculation, we use \emph{Manhattan distance} as a heuristic which is a  lower bound of the actual path cost. Therefore, the heuristic is admissible, and our algorithm generates the shortest edge costs.

The $\mathtt{plan\_in\_H}$ procedure, which is invoked on line~\ref{plan_in_H}, checks if a plan exists from current robot position $pos_{cur}$ at $time_{cur}$ within the horizon $H$. This procedure firstly generates the constraints for modeling the decision problem for the robot from the current timestamp using  $\mathtt{gen\_cons}$ procedure.
The procedure $\mathtt{gen\_cons}$ on line~\ref{get_cons} runs Dijkstra's algorithm on $G_r$ to get minimum length prefix path or suffix cycle using the pre-computed edge costs from $\mathtt{dy\_cost}$ method. The generated constraints are then given to an SMT solver (line~\ref{solve_cons}). 
Procedure $\mathtt{get\_plan\_from\_model}$ on line~\ref{get_plan} parses the model solution to get path within $H$ from optimal decision sequence. 
The upper bound on the sum of the the durations  to execute the procedures $\mathtt{dy\_cost}$ and $\mathtt{plan\_in\_H}$ is referred to as $time_{comp}$.  

The $\mathtt{plan\_in\_H}$ procedure returns the optimal decision sequence within the given horizon $H$ from the current robot position $pos_{cur}$ at time $time_{cur}$, if one exists.
If it does not, then it indicates that from current position $pos_{cur}$ between time $time_{cur}$ to $time_{cur}+H$, not even one cycle could be completed. 
In this case, we re-plan for the time window from $time_{cur} + time_{comp}$ to $time_{cur}+time_{comp}+H$.
A solution having at least one cycle could exist in this duration because the unavailability time of grid cells decreases as time elapses. 
We repeat this step until we get an optimal decision sequence from current robot position (line~\ref{update_time}-\ref{plan_in_H}).
\noindent 
\longversion
{
Figure~\ref{fig3} shows the block diagram for $\textsf{DT*}$.
} 

\begin{figure}
  \centering
   \includegraphics[height=4.5cm,width=8.5cm]{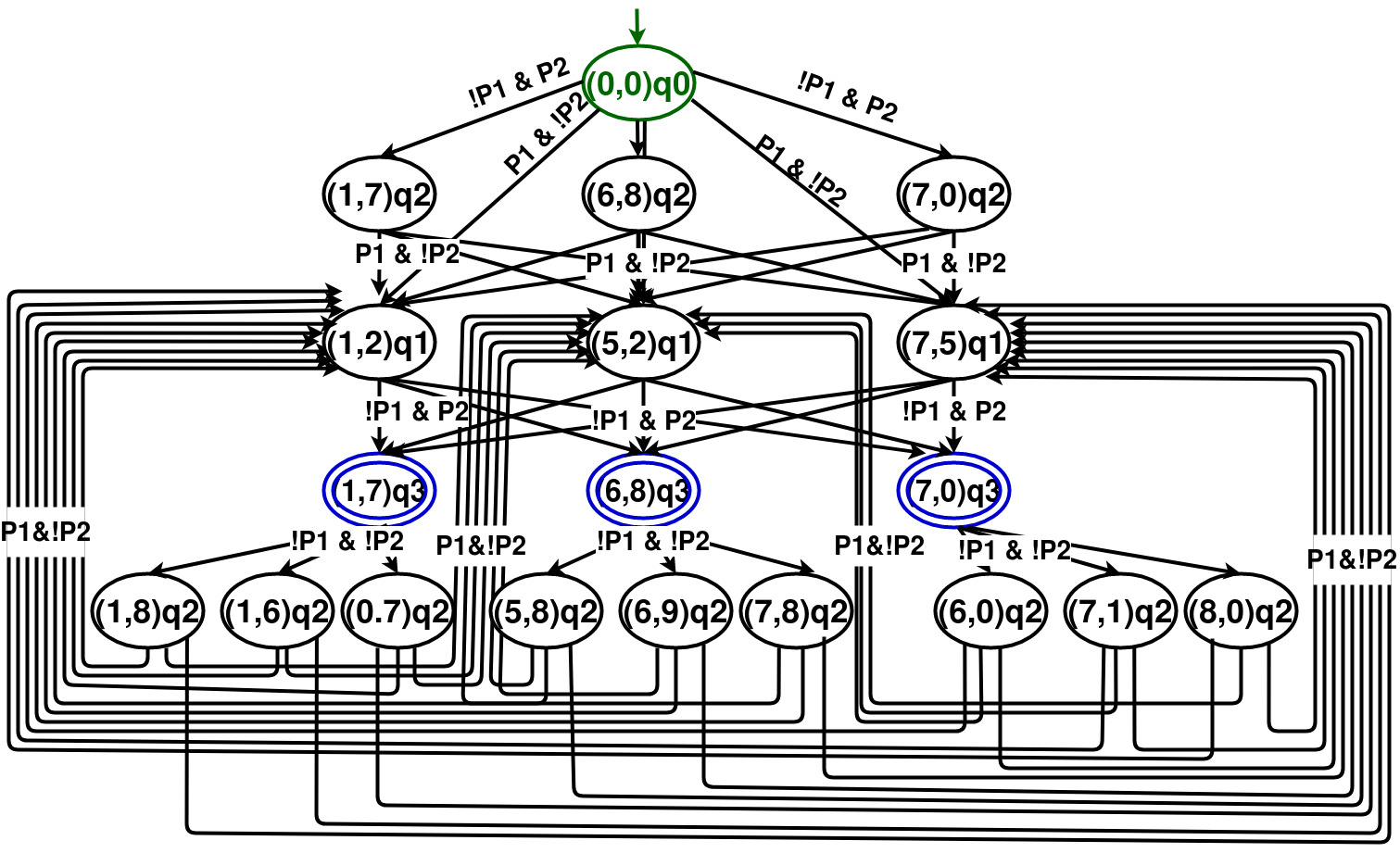}
      \includegraphics[height=4.2cm,width=8.5cm]{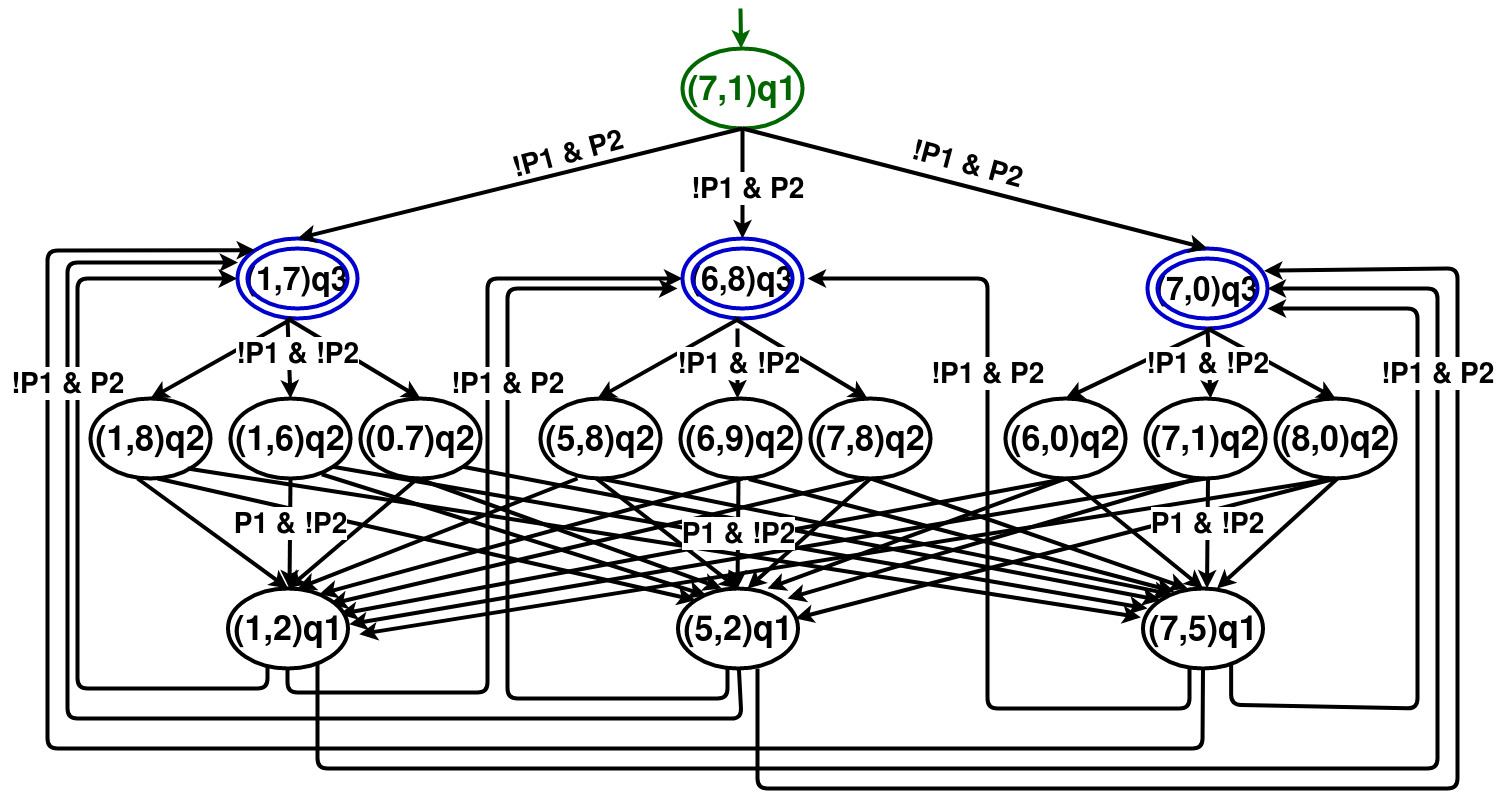}
   \caption{(a) Initial product graph for workspace given in Figure~\ref{fig1}(b), (b) Updated product graph $G_r$ based on environment changes at $ts=10$ shown in Figure~\ref{fig1}(c)}
   \label{fig4}
\end{figure}

\begin{example}
Figure~\ref{fig4} shows the change(s) in $G_r$ for the example that we have discussed earlier in section ~\ref{sec-example}. Figure~\ref{fig4}(a) shows the reduced product graph $G_r$ in the initial static environment. When the environment changes at timestamp 10, we update the previous graph $G_r$ as given on line~\ref{atg} based on the current robot position and its B\"{u}chi state. The updated $G_r$ is shown in Figure~\ref{fig4}(b). The dynamic information captured on line~\ref{env_changes} for our example will be as follows.
Any action $act\in Action$ from the neighbouring grid cells of $\mathtt{P3}$ to $\mathtt{P3}$, on time $t\in(10,25)$ will cost $25-t+1$. And any $act\in Action$ from $\mathtt{P3}$ to the neighbouring cells of $\mathtt{P3}$ on time $t\in(10,25)$ cannot be taken as the location is blocked. Similar will be the case with proposition location $\mathtt{P1}$.
This information will then be passed to $\mathtt{dy\_cost}$ method to get edge cost of $G_r$ within $H$. The plan generated by the $\mathtt{plan\_in\_H}$ method is shown as a timeline in figure~\ref{fig2}(c).
\end{example}

\longversion{
\begin{figure}
  \centering
  \includegraphics[width=\linewidth]{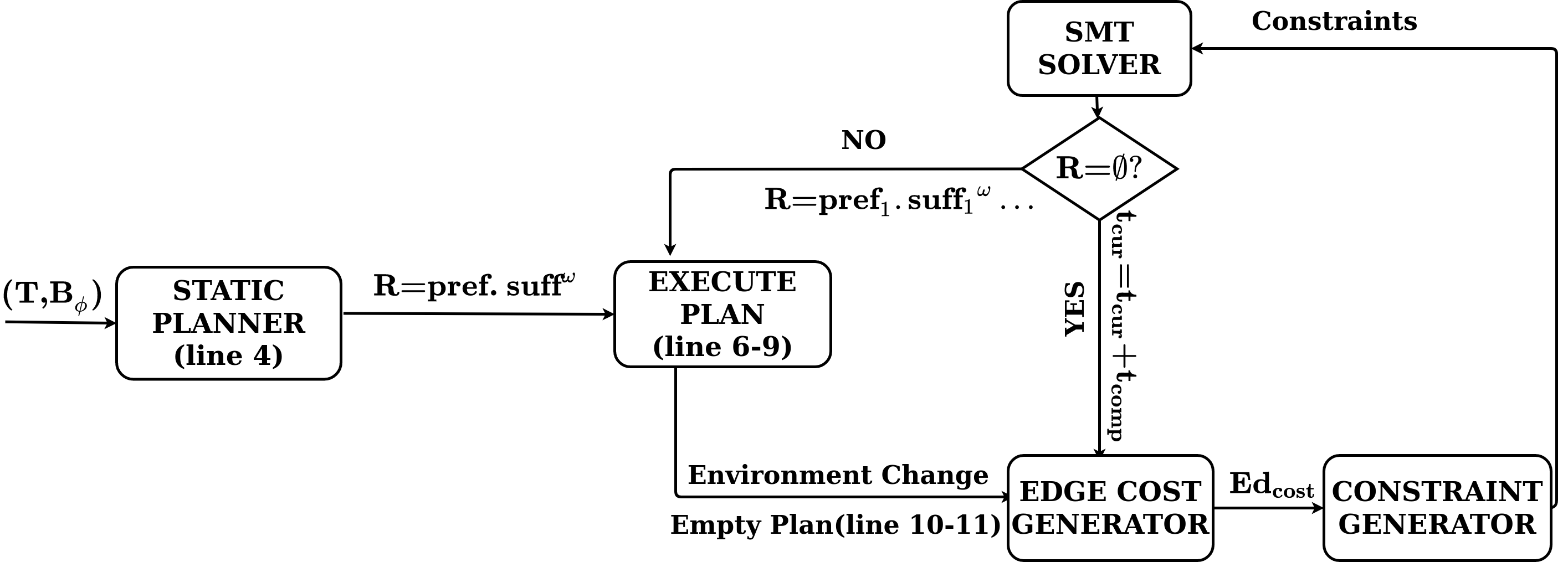}
  \caption{Block diagram for $\textsf{DT*}$}
  \label{fig3}
\end{figure}
}
\subsection{Generating Optimization Model}
In this section, we define the variables, constraints, and objective functions for the optimization problem.

 \comment{
  \begin{lemma}
  The best choice for the location at which decision to switch the cycle is to be made $\in F_r$  
  \begin{proof}
  Let's, take an example where the cycle completion requires the propositions to be visited in  ${(p_1,p_2,p_3,p1)}^\omega$ order\\
  Once, cycle completes at location $l_1$, where $p_1$ is true, the robot decides to switch to some other cycle $cycle_2$ by choosing path $path_1$ at time $t$\\
  Or, the robot can go from $l_1$, to some other location $l_2$, then to $l_3$, where $p_2,p_3$ are true true respectively, by taking path $path_2$, then from $l_3$ it switches to $cycle_2$, by taking path $path_3$.
  Then, $time(path_1)\leq time(path_2+path_3)$\\
  This follows from the fact that we are running dijkstra's algorithm on $l_1$, which will give minimum length path from $l_1$, so $time(path_1)$ provides the lower bound on the time which is required to switch cycle from $l_1$. So, any other path can either be of equal or greater length.\\
  Hence, the best location from which the robot must decide to switch the cycle must be $\in F_r$
  
  \end{proof}
  \end{lemma}
  }

\medskip
\noindent
\textbf{Decision variables:}  Let us denote the state $(x_i,y_i,s_i)$ of $G_r$ by $l_i$ 
, where $(x_i,y_i)$ is a 2D-grid coordinate and $s_i$ is a state in the B\"{u}chi automata.
We define the Boolean variable $X_{l_it_i}$ which becomes $\mathtt{true}$ iff the robot is at location $l_i$ at time $t_i$.  
We define the Boolean variable $C_{l_it_i\tau_i}$ which becomes $\mathtt{true}$ iff 
the robot completes 
a cycle of length $\tau_i$ at time $t_i$ from location $l_i$. Here, $l_i \in F_r$ and $\tau_i$ is the shortest cycle length among those cycles starting at $l_i$ at time $t_i - \tau_i$.



      
  
 
For every decision that could be taken at any location $l_i$ at some time $t_i$ to reach location $l_j$ at $t_j$, we define a Boolean variable $A_{l_il_jt_i}$ which captures if the decision to move from $l_i$ to $l_j$ was taken by the robot at time $t_i$. 
Here, $t_j-t_i$ denotes the time taken to cover a cycle or a prefix path. Thus, $A_{l_il_jt_i}$ is $\mathtt{true}$ iff both $X_{l_it_i}$ and $X_{l_jt_j}$ are $\mathtt{true}$. 
Also, we define the Boolean variable $B_{t_it_j}$ which is $\mathtt{true}$ iff there exists a $t_k$,  $t_i< t_k< t_j$, such that some $X_{l_kt_k} = \mathtt{true}$.




\begin{figure}
    \centering
    \includegraphics[width=0.95\linewidth,height=3.2cm]{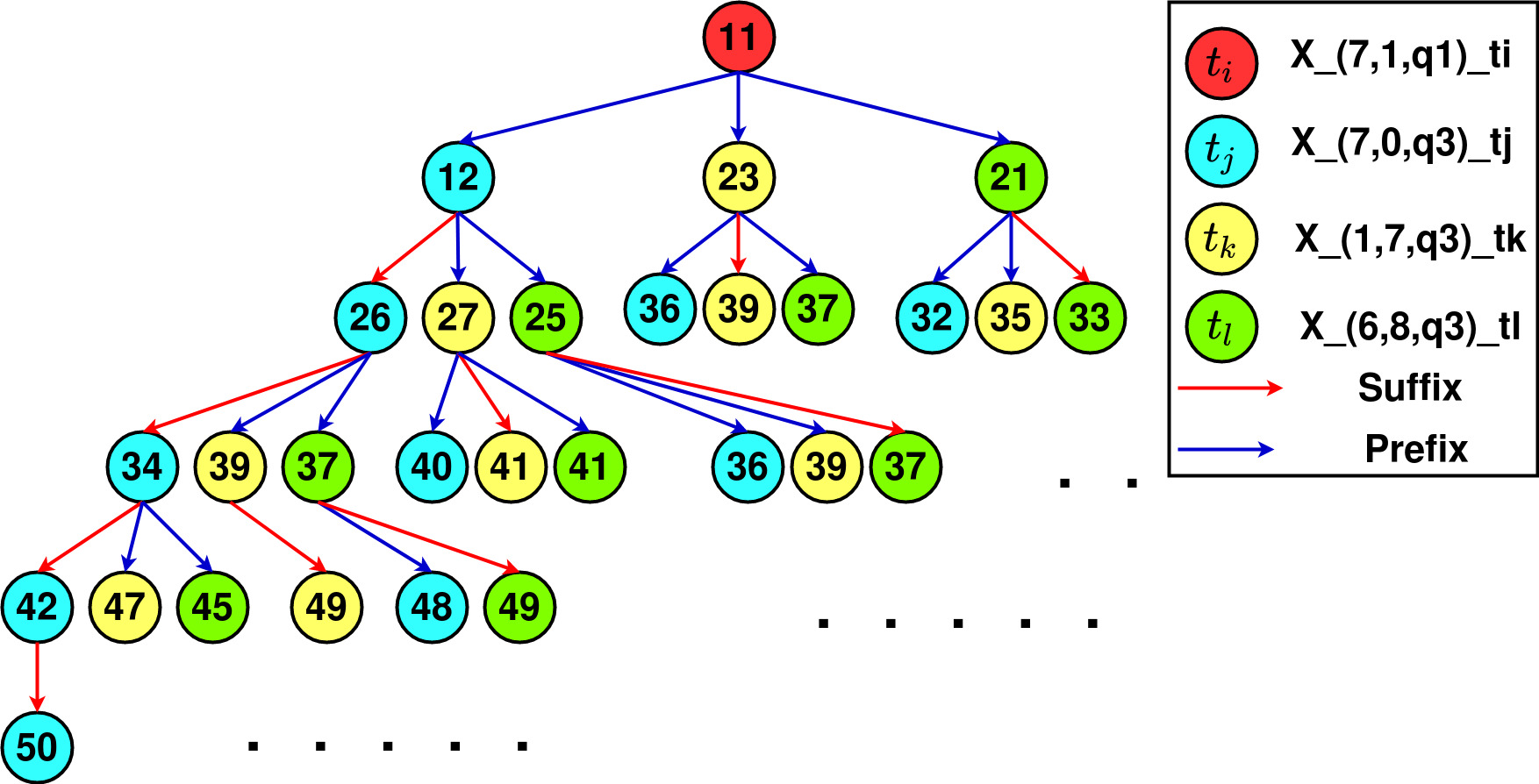}
    \caption{Generation of model constraints for the example given in Figure~\ref{fig1}(c)}
    \label{fig5}
\end{figure}
\medskip
\noindent
\textbf{Constraints:}
We  present  the constrains in the optimization problem below.

\smallskip
\noindent
\textit{1) Movement between locations:}
Let $t_0$ be the time at which we have to re-plan, and $l_0$ be the current location of the robot. Then we can write
\begin{equation}
X_{l_0t_0} \Longleftrightarrow \mathtt{true}.
 \end{equation}
 If at time $t_i$, the robot is at location $l_i$, then the next location has to be decided from the set of nodes reachable from $l_i$ in $G_r$.
The set of decisions that could be made at location $l_i$ are: covering various prefixes starting from $l_i$ and reaching destination nodes. If $l_i$ itself is a destination node, then going to $l_i$ denotes completion of a cycle. Suppose, $l_1 \ldots l_n$ be the set of destination nodes that could be reached by covering various prefixes from $l_i$ at time $t_i$.
 To capture this decision in our model using decision variables, we write:
 \begin{equation}
  X_{l_it_i} \implies \ \bigvee_{j=1}^n X_{l_jt_j}
  \label{motion_eq}
 \end{equation}
\begin{equation}
\bigwedge_{j=1}^n \big(X_{l_jt_j} \implies \bigwedge_{k \in \{1, \ldots, n\} \setminus j} \neg X_{l_kt_k} \big)
\label{sos_eq}
 \end{equation}
Here, $ t_1,t_2,\ldots, t_n$ are the times of completion of a prefix path or suffix cycle from $l_i$.

Figure~\ref{fig5} shows the expansion of the decision tree for the example shown in Figure~\ref{fig1}. The tree shows the choice of possible decisions that the robot could take at any location. From the initial location $(7,1,q_1)$, the robot can traverse 
three different prefixes 
to reach the destination locations $(7,0,q_3),(1,7,q_3),(6,8,q_3)$ respectively and so on. The blue and red edges in this tree denote the decision of taking prefix and suffix respectively, and the node values denote the time of completion.
 
\smallskip
\noindent
\textit{2) Cycle completion constraints:}
If location $l_i$ was a destination node in the reduced graph, and if a cycle could be completed within the horizon length, then we can add a constraint to capture the decision of completing a cycle from $l_i$.
So along with constraints~\eqref{motion_eq} and ~\eqref{sos_eq}, we also need to capture this completion of a cycle using cycle completion variables:
\begin{equation}
 C_{l_it_j(t_j-t_i)} \Longleftrightarrow  X_{l_it_i} \wedge X_{l_it_j}
\end{equation}
Here, $(t_j-t_i)$ is the time taken to complete a cycle from location $l_i$ at time $t_i$.
 The above constraints says that $  C_{l_it_j(t_j-t_i)}$  will be set to $\mathtt{true}$ iff the robot was at destination location $l_i$ at time $t_i$, where it took a cycle and reached $l_i$ again at time $t_j$.

\smallskip
\noindent
\textit{3) Integrity Constraints:}
This constraint ensures that the robot cannot be at multiple locations at the same timestamp.
Let $\{l_1,l_2,\ldots, l_m\}$ be the set of all possible reachable locations at some timestamp $t$. So, we add the following integrity constraint:
\begin{equation}
\bigwedge_{i=1}^m \big(X_{l_it} \implies \bigwedge_{j \in \{1, \ldots, m\} \setminus i} \neg X_{l_jt}\big)
\label{int_eq}
\end{equation}

\smallskip
\noindent
\textit{4) Continuity Constraints:}
We need to ensure that if the robot is at location $l_i$ at time $t_i$ where it took a decision to reach $l_j$ at $t_j$ such that $t_j-t_i$ is the path (or cycle) cost from $l_i$ to $l_j$ ($A_{l_il_jt_i}$ is $\mathtt{true}$), then any other decision could not be taken at any intermediate timestamp $t_k$ s.t. $t_i < t_k< t_j$ ($B_{t_it_j}$ is $\mathtt{false}$).
The following constraint captures the same.
\begin{equation}
\neg (A_{l_il_jt_i} \ \wedge \  B_{t_it_j})
\label{cont_eq}
\end{equation}


\medskip
\noindent
\textbf{Objective Function:}
The objective function of the optimization problem contains the following three objectives.





\smallskip
\noindent
\textit{1) Maximize the number of cycles that can be completed within the horizon $H$.} 
Here the objective is to maximize the number of all the cycle variables, which are set to $\mathtt{true}$. 

\longversion{
We introduce an integer variable $cy_{count}$ which gets incremented when any $C_{l_it_i\tau_i}$ variable is set to  $\mathtt{true}$ for all $l_i\in F_r$  and $time_{cur}\leq t_i \leq time_{cur}+H$.

Now the objective is: $\mathtt{Maximize} \ cy_{count}$.
}

\smallskip
\noindent
\textit{2) Minimize the length of the last cycle.} Suppose the maximum number of suffix cycles that could be covered within $H$ is $k$. Our secondary objective is then to choose that solution that has the minimum cost for the  $k$-{th} cycle out of all possible solutions. This is because we want the robot to stay close to the shortest length cycle when the robot finishes traversing the generated plan. 

\longversion{
We order all the cycle completion variables $C_{l_it_i\tau_i}$ with decreasing completion times, i.e., $t_i$. Let the timestamps ordered by decreasing completion time be : $t_{n_m}$, $t_{n_{m-1}} \ldots$. 
One must note that there could be multiple decision sequences that could result in the completion of the $k$-{th} cycle at any timestamp $t_{n_{m-i}}$. 
Let the number of $C_{l_it_{n_{m-i}}\tau_i}$ variables that could result in completion of $k^{th}$ cycle at any timestamp $t_{n_{m-i}}$ be $o$.
Now, we define an integer variable $last_{len}(n_m)$ to capture the length of the last cycle completed based on the current model solution. The recursive definition to calculate $last_{len}$'s value is outlined below:
\begin{align}
last\_&len(n_m) = (\bigvee_{i=1}^o C_{l_it_{n_m}\tau_i})? \ \ \ \ \nonumber \\ & \ (\tau_i \ | \ C_{l_it_{n_m}\tau_i}, i \in \{1, \ldots, o\}): last\_len(n_{m-1})
\end{align}
The objective can be written as:  $\mathtt{Minimize}: last\_len$.
}

\smallskip
\noindent
\textit{3) Minimize the time at which the last cycle is completed within the horizon.}
This objective is important because we want to ensure that the plan generated by the solver is of minimal length such that $k$ cycles get completed in the minimum possible time within the horizon $H$.

\longversion{
We again define an integer variable $T_{total}$, which denotes the time at which the last cycle was completed. Similar to the procedure given for calculating $last_{len}$, we can calculate this variable recursively starting from the last completion time.
\begin{equation}
T_{total}(n_m) = (\bigvee_{i=1}^k C_{l_it_{n_m}\tau_i})? t_{n_m} : T_{total}(n_{m-1})
\end{equation}
The objective is: $\mathtt{Minimize}: T_{total}$.
}

\shortversion{
The formal definitions of the objective functions are provided in the full version~\cite{PurohitS21}.
}
The first objective function is assigned the highest priority, and the third one the lowest priority. In case there are multiple solutions with the same value for the first objective function, the second objective function is applied. The third objective function is utilized to distinguish between the solutions having the same value for the first two objective functions.



\subsection{Theoretical Guarantees}

$\textsf{DT*}$ provides the following theoretical guarantees. 
\shortversion{
The proofs of the theorems are available in the extended version of the paper~\cite{PurohitS21}.
}

\begin{theorem} [\textbf{Soundness}]
 The plan generated by $\textsf{DT*}$ always obeys the LTL specification $\phi$.
\end{theorem}
\longversion {
\begin{proof}
The plan generated by $\textsf{DT*}$ is of the form of a sequence of decisions with their timestamps at which the decisions need to be implemented.
Say the solver generated decision sequence is of the form:
  $\langle (l_1,t_1),(l_2,t_2)(l_3,t_3)\ldots(l_k,t_k)\rangle$.
  
$\mathbf{(i)}$ If $l_i=l_{i+1}$ for any $i \in \{1, \ldots, k\}$ in the above solution, then it denotes that the robot must traverse a cycle  of length $t_{i+1} - t_i$ from $l_i$ at time $t_i$. The length of the cycle is calculated by running Dijkstra's algorithm on $G_r$ whose edge costs are calculated by using heuristic based approach. 
The graph $G_r$ differs from the product graph $P$ for the nodes which are not adjacent in $P$, but obey the distinct node condition, as mentioned in~\cite{Tstar}. 
Thus, the path obeys the constraints imposed by the LTL specification $\phi$. 
If the nodes are adjacent in $P$, then they obey $\phi$ by the product graph definition.
And hence, entire cycle from $l_i$ obeys $\phi$.
  
$\mathbf{(ii)}$ Any $l_i \neq l_{i+1}$ in the above solution denotes a prefix path from $l_i$ to $l_{i+1}$, which also obeys $\phi$ using the same reasoning.
\end{proof}
}

This theorem ensures that our proposed algorithm provides the first guarantee required in solving Problem 1.

\begin{theorem} [\textbf{Optimality within horizon}]
  Given a start location and a horizon length $H$, the plan generated by $\textsf{DT*}$ is optimal in terms of the number of cycles completed within the horizon.
\end{theorem}

\longversion {
\begin{proof}
  As per the constraints, $\textsf{DT*}$ does not discard any cycle within horizon $H$ from the current position of the robot. The search space covers all possible cycles that could be traversed in some time within $H$. The solution generated by the solver maximizes the primary objective, which leads to the maximization of the number of cycles that could be covered within $H$ from the current robot location. The other two objectives are secondary, and they do not have any adverse effect on the primary objective.
Hence, from a fixed start location and within the horizon length $H$, the solver always maximizes the number of cycles that could be taken. 
\end{proof}
}
This result is established keeping the second required guarantee in Problem 1 in mind, which requires that the trajectory up to any time point covers the maximum possible number of loops. Though the above theorem does not provide this guarantee, it attempts to keep the number of traversed suffixes maximum in each horizon when the plan is computed. This is the best we can achieve in a receding horizon planning setting.
The possibility of achieving the global optimality is ruled out as the knowledge of the dynamic obstacles is not known beforehand and becomes available during the operation.
\shortversion{
In the full version~\cite{PurohitS21}, we provide an example to show how a greedy algorithm could outperform $\textsf{DT*}$ in a long duration. Our experimental results establish that such instances are rare, and overall $\textsf{DT*}$ offers superior performance than both the greedy algorithms.
}

\longversion{
\begin{figure}[t]
\begin{subfigure}{{\textwidth}}
\begin{tikzpicture}[%
    every node/.style={
        font=\scriptsize,
        text height=1ex,
        text depth=.25ex,
    },
]
\draw[->] (0,0) -- (8.0,0);
\foreach \x in {0,0.83,...,8}{
    \draw (\x cm,1.5pt) -- (\x cm,0pt);
}
\node[anchor=north] at (0,0) {$3$};
\node[anchor=north] at (1.6,0) {$19$};
\node[anchor=north] at (2.3,0) {$27$};
\node[anchor=north] at (3,0) {$35$};
\node[anchor=north] at (3.7,0) {$43$};
\node[anchor=north] at (4.4,0) {$51$};
\node[anchor=north] at (5.1,0) {$59$};
\node[anchor=north] at (5.8,0) {$67$};
\node[anchor=north] at (6.5,0) {$75$};
\node[anchor=north] at (7.2,0) {$83$};

\node[anchor=south] at (0,0) {$(1,2)$};
\node[anchor=south] at (1.6,0) {$(7,0)$};
\node[anchor=south] at (2.3,0) {$(7,0)$};
\node[anchor=south] at (3,0) {$(7,0)$};
\node[anchor=south] at (3.7,0) {$(7,0)$};
\node[anchor=south] at (4.4,0) {$(7,0)$};
\node[anchor=south] at (5.1,0) {$(7,0)$};
\node[anchor=south] at (5.8,0) {$(7,0)$};
\node[anchor=south] at (6.5,0) {$(7,0)$};
\node[anchor=south] at (7.2,0) {$(7,0)$};
\node[anchor=north] at (7.8,0) {Time};
\node[anchor=south] at (7.8,0) {Loc};
\draw[dg,thick,-latex] (0,-0.5) -- (1.6,-0.5);

\fill[dg] (1.6,-0.4) rectangle (1.95,-0.55);
\fill[g] (1.95,-0.4) rectangle (2.3,-0.55);
\fill[dg] (2.3,-0.4) rectangle (2.65,-0.55);
\fill[g] (2.65,-0.4) rectangle (3,-0.55);
\fill[dg] (3,-0.4) rectangle (3.35,-0.55);
\fill[g] (3.35,-0.4) rectangle (3.7,-0.55);
\fill[dg] (3.7,-0.4) rectangle (4.05,-0.55);
\fill[g] (4.05,-0.4) rectangle (4.4,-0.55);
\fill[dg] (4.4,-0.4) rectangle (4.85,-0.55);
\fill[g] (4.85,-0.4) rectangle (5.1,-0.55);
\fill[dg] (5.1,-0.4) rectangle (5.45,-0.55);
\fill[g] (5.45,-0.4) rectangle (5.8,-0.55);
\fill[dg] (5.8,-0.4) rectangle (6.15,-0.55);
\fill[g] (6.15,-0.4) rectangle (6.5,-0.55);
\fill[dg] (6.5,-0.4) rectangle (6.85,-0.55);
\fill[g] (6.85,-0.4) rectangle (7.2,-0.55);
\end{tikzpicture}
\caption{\ \ \ \ \ \ \ \ \ \ \ \ \ \ \ \ \ \ \ \ \ \ \ \ \ \ \ \ \ \ \ \ \ \ \ \ \ \ \ \ \ \ \ \ \ \ \ \ \ \ \ \ \ \ \ \ \ \ \ \ \ \ \ \ \ \ \ \ \ \ \ \ \ \ \ \ \ \ \ \ \ \ \ \ }
\end{subfigure}

\begin{subfigure}{{\textwidth}}
\begin{tikzpicture}[%
    every node/.style={
        font=\scriptsize,
        text height=1ex,
        text depth=.25ex,
    },
]
\draw[->] (0,0) -- (8.0,0);
\foreach \x in {0,0.83,...,8}{
    \draw (\x cm,1.5pt) -- (\x cm,0pt);
}
\node[anchor=north] at (0,0) {$3$};
\node[anchor=north] at (0.7,0) {$8$};
\node[anchor=north] at (1.6,0) {$18$};
\node[anchor=north] at (2.5,0) {$28$};
\node[anchor=north] at (3.4,0) {$38$};
\node[anchor=north] at (4.3,0) {$48$};
\node[anchor=north] at (5.2,0) {$58$};
\node[anchor=north] at (6.1,0) {$68$};
\node[anchor=north] at (7.0,0) {$78$};

\node[anchor=south] at (0,0) {$(1,2)$};
\node[anchor=south] at (0.7,0) {$(1,7)$};
\node[anchor=south] at (1.6,0) {$(1,7)$};
\node[anchor=south] at (2.5,0) {$(1,7)$};
\node[anchor=south] at (3.4,0) {$(1,7)$};
\node[anchor=south] at (4.3,0) {$(1,7)$};
\node[anchor=south] at (5.2,0) {$(1,7)$};
\node[anchor=south] at (6.1,0) {$(1,7)$};
\node[anchor=south] at (7.0,0) {$(1,7)$};
\node[anchor=north] at (7.8,0) {Time};
\node[anchor=south] at (7.8,0) {Loc};
\draw[dg,thick,-latex] (0,-0.5) -- (0.7,-0.5);

\fill[dg] (0.7,-0.4) rectangle (1.15,-0.55);
\fill[g] (1.15,-0.4) rectangle (1.6,-0.55);
\fill[dg] (1.6,-0.4) rectangle (2.05,-0.55);
\fill[g] (2.05,-0.4) rectangle (2.5,-0.55);
\fill[dg] (2.5,-0.4) rectangle (2.95,-0.55);
\fill[g] (2.95,-0.4) rectangle (3.4,-0.55);
\fill[dg] (3.4,-0.4) rectangle (3.85,-0.55);
\fill[g] (3.85,-0.4) rectangle (4.3,-0.55);
\fill[dg] (4.3,-0.4) rectangle (4.75,-0.55);;
\fill[g] (4.75,-0.4) rectangle (5.2,-0.55);
\fill[dg] (5.2,-0.4) rectangle (5.65,-0.55);
\fill[g] (5.65,-0.4) rectangle (6.1,-0.55);
\fill[dg] (6.1,-0.4) rectangle (6.55,-0.55);
\fill[g] (6.55,-0.4) rectangle (7,-0.55);

\end{tikzpicture}
\caption{\ \ \ \ \ \ \ \ \ \ \ \ \ \ \ \ \ \ \ \ \ \ \ \ \ \ \ \ \ \ \ \ \ \ \ \ \ \ \ \ \ \ \ \ \ \ \ \ \ \ \ \ \ \ \ \ \ \ \ \ \ \ \ \ \ \ \ \ \ \ \ \ \ \ \ \ \ \ \ \ \ \ \ \ }
\end{subfigure}

\begin{subfigure}{{\textwidth}}
\begin{tikzpicture}[%
    every node/.style={
        font=\scriptsize,
        text height=1ex,
        text depth=.25ex,
    },
]
\draw[->] (0,0) -- (8.0,0);
\foreach \x in {0,0.83,...,8}{
    \draw (\x cm,1.5pt) -- (\x cm,0pt);
}
\node[anchor=north] at (0,0) {$3$};
\node[anchor=north] at (0.7,0) {$9$};
\node[anchor=north] at (1.6,0) {$19$};
\node[anchor=north] at (2.5,0) {$29$};
\node[anchor=north] at (3.7,0) {$43$};
\node[anchor=north] at (4.4,0) {$51$};
\node[anchor=north] at (5.1,0) {$59$};
\node[anchor=north] at (5.8,0) {$68$};
\node[anchor=north] at (6.6,0) {$76$};
\node[anchor=north] at (7.3,0) {$84$};

\node[anchor=south] at (0,0) {$(1,2)$};
\node[anchor=south] at (0.7,0) {$(1,7)$};
\node[anchor=south] at (1.6,0) {$(1,7)$};
\node[anchor=south] at (2.5,0) {$(1,7)$};
\node[anchor=south] at (3.7,0) {$(7,0)$};
\node[anchor=south] at (4.4,0) {$(7,0)$};
\node[anchor=south] at (5.1,0) {$(7,0)$};
\node[anchor=south] at (5.8,0) {$(7,0)$};
\node[anchor=south] at (6.6,0) {$(7,0)$};
\node[anchor=south] at (7.3,0) {$(7,0)$};
\node[anchor=north] at (7.9,0) {Time};
\node[anchor=south] at (7.9,0) {Loc};
\draw[dg,thick,-latex] (0,-0.5) -- (0.7,-0.5);

\fill[dg] (0.7,-0.4) rectangle (1.15,-0.55);
\fill[g] (1.15,-0.4) rectangle (1.6,-0.55);
\fill[dg] (1.6,-0.4) rectangle (2.05,-0.55);
\fill[g] (2.05,-0.4) rectangle (2.5,-0.55);
\draw[dg,thick,-latex] (2.6,-0.5) -- (3.7,-0.5);
\fill[dg] (3.7,-0.4) rectangle (4.05,-0.55);
\fill[g] (4.05,-0.4) rectangle (4.4,-0.55);
\fill[dg] (4.4,-0.4) rectangle (4.75,-0.55);
\fill[g] (4.75,-0.4) rectangle (5.1,-0.55);
\fill[dg] (5.2,-0.4) rectangle (5.55,-0.55);
\fill[g] (5.55,-0.4) rectangle (5.9,-0.55);
\fill[dg] (5.9,-0.4) rectangle (6.25,-0.55);
\fill[g] (6.25,-0.4) rectangle (6.6,-0.55);
\fill[dg] (6.6,-0.4) rectangle (6.95,-0.55);
\fill[g] (6.95,-0.4) rectangle (7.3,-0.55);

\draw[decorate,decoration={brace,amplitude=5pt}] (0.1,0.45) -- (2.5,0.45)
    node[anchor=south,midway,above=4pt] {Plan1};

\draw[decorate,decoration={brace,amplitude=5pt}] (2.6,0.45) -- (5.1,0.45)
    node[anchor=south,midway,above=4pt] {Plan2};
    
\draw[decorate,decoration={brace,amplitude=5pt}] (5.2,0.45) -- (7.3,0.45)
    node[anchor=south,midway,above=4pt] {Plan3};

\end{tikzpicture}
\caption{\ \ \ \ \ \ \ \ \ \ \ \ \ \ \ \ \ \ \ \ \ \ \ \ \ \ \ \ \ \ \ \ \ \ \ \ \ \ \ \ \ \ \ \ \ \ \ \ \ \ \ \ \ \ \ \ \ \ \ \ \ \ \ \ \ \ \ \ \ \ \ \ \ \ \ \ \ \ \ \ \ \ \ \ }
\end{subfigure}
\begin{subfigure}{{\textwidth}}
\begin{tikzpicture}[%
    every node/.style={
        font=\scriptsize,
        text height=1ex,
        text depth=.25ex,
    },
]
\draw[->] (0,0) -- (8.0,0);
\foreach \x in {0,0.83,...,8}{
    \draw (\x cm,1.5pt) -- (\x cm,0pt);
}
\node[anchor=north] at (0,0) {$3$};
\node[anchor=north] at (0.8,0) {$10$};
\node[anchor=north] at (1.6,0) {$19$};
\node[anchor=north] at (2.3,0) {$27$};
\node[anchor=north] at (3.0,0) {$35$};
\node[anchor=north] at (3.7,0) {$43$};
\node[anchor=north] at (4.4,0) {$51$};
\node[anchor=north] at (5.2,0) {$60$};
\node[anchor=north] at (5.9,0) {$68$};
\node[anchor=north] at (6.6,0) {$76$};
\node[anchor=north] at (7.3,0) {$84$};

\node[anchor=south] at (0,0) {$(1,2)$};
\node[anchor=south] at (1.6,0) {$(7,0)$};
\node[anchor=south] at (2.3,0) {$(7,0)$};
\node[anchor=south] at (3,0) {$(7,0)$};
\node[anchor=south] at (3.7,0) {$(7,0)$};
\node[anchor=south] at (4.4,0) {$(7,0)$};
\node[anchor=south] at (5.2,0) {$(7,0)$};
\node[anchor=south] at (5.9,0) {$(7,0)$};
\node[anchor=south] at (6.6,0) {$(7,0)$};
\node[anchor=south] at (7.3,0) {$(7,0)$};
\node[anchor=north] at (7.9,0) {Time};
\node[anchor=south] at (7.9,0) {Loc};
\draw[dg,thick,-latex] (0,-0.5) -- (1.6,-0.5);

\fill[dg] (1.6,-0.4) rectangle (1.95,-0.55);
\fill[g] (1.95,-0.4) rectangle (2.3,-0.55);
\fill[dg] (2.3,-0.4) rectangle (2.65,-0.55);
\fill[g] (2.65,-0.4) rectangle (3,-0.55);
\fill[dg] (3,-0.4) rectangle (3.35,-0.55);
\fill[g] (3.35,-0.4) rectangle (3.7,-0.55);
\fill[dg] (3.7,-0.4) rectangle (4.05,-0.55);
\fill[g] (4.05,-0.4) rectangle (4.4,-0.55);
\fill[dg] (4.5,-0.4) rectangle (4.85,-0.55);
\fill[g] (4.85,-0.4) rectangle (5.2,-0.55);
\fill[dg] (5.2,-0.4) rectangle (5.55,-0.55);
\fill[g] (5.55,-0.4) rectangle (5.9,-0.55);
\fill[dg] (5.9,-0.4) rectangle (6.25,-0.55);
\fill[g] (6.25,-0.4) rectangle (6.6,-0.55);
\fill[dg] (6.6,-0.4) rectangle (6.95,-0.55);
\fill[g] (6.95,-0.4) rectangle (7.3,-0.55);
\draw[decorate,decoration={brace,amplitude=5pt}] (0.1,0.45) -- (4.4,0.45)
    node[anchor=south,midway,above=4pt] {Plan1};

\draw[decorate,decoration={brace,amplitude=5pt}] (4.5,0.45) -- (7.3,0.45)
    node[anchor=south,midway,above=4pt] {Plan2};
\end{tikzpicture}
\caption{\ \ \ \ \ \ \ \ \ \ \ \ \ \ \ \ \ \ \ \ \ \ \ \ \ \ \ \ \ \ \ \ \ \ \ \ \ \ \ \ \ \ \ \ \ \ \ \ \ \ \ \ \ \ \ \ \ \ \ \ \ \ \ \ \ \ \ \ \ \ \ \ \ \ \ \ \ \ \ \ \ \ \ \ }
\end{subfigure}

\caption{Timelines for the plan generated by (a) \greedyone Algorithm (b) \greedytwo Algorithm (c) $\textsf{DT*}$ with $H=29$ (d) $\textsf{DT*}$ with $H=49$}
\label{timeline2}
\end{figure}
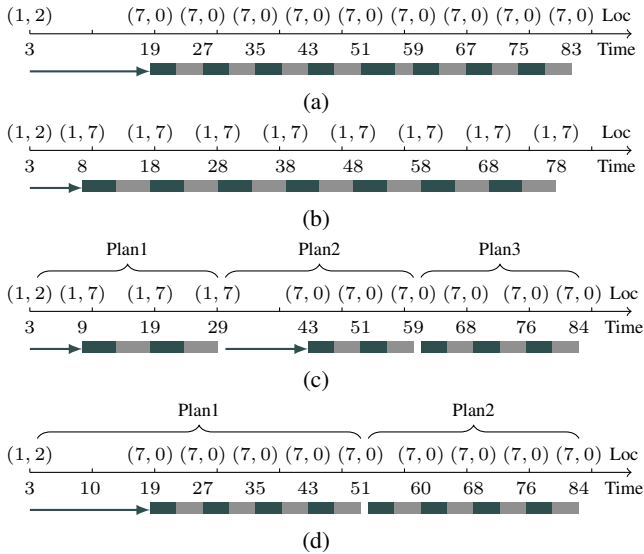

\begin{example}
Let us consider an example to understand how $\greedyone$ algorithm could outperform $\textsf{DT*}$ by enabling the robot to cover more number of cycles within a fixed time duration in some situations. For the workspace described in section~\ref{sec-example}, assume that the dynamic change occurs at timestamp 3 and proposition location $D3$ becomes unavailable till timestamp 18.

$\greedyone$ decides to wait for $D3$ to become available. This choice incurs a larger prefix path but $\greedyone$ still chooses it as $\langle (7,0),(5,2),(7,0)\rangle$ is the minimum length cycle in the current environment. As shown in Figure~\ref{timeline2}(a), till total planning time of $85$, the number of cycles that could be covered by $\greedyone$ algorithm is $8$.

Solution generated by $\greedytwo$ algorithm is shown in Figure~\ref{timeline2}(b). As $\greedytwo$ algorithm minimizes the time for completion of just one cycle, it chooses $\langle (1,7),(1,2),(1,7)\rangle$ cycle which has a smaller prefix but a larger corresponding suffix cycle. The total number of cycles that could be covered by $\greedytwo$ algorithm's plan till timestamp $85$ is $7$.

Suppose, for $\textsf{DT*}$, we choose the horizon $H$ to be $29$. $\textsf{DT*}$ starts planning from timestamp $4$ as it incurs $1\si{\second}$ to generate the plan by solving the constraints. 
It could cover $2$ cycles if it chose $\langle (1,7),(1,2),(1,7)\rangle$ cycle at timestamps $19$ and $29$ respectively. Whereas, if it waits for 
$D3$ to become available then it could cover only one cycle at $27$. As shown in Figure~\ref{timeline2}(c), $\textsf{DT*}$ chooses cycle $\langle (1,7),(1,2),(1,7)\rangle$ to maximize the number of cycles covered within horizon $H=29$.
At timestamp $29$ replanning happens. The choices available now are to cover two cycles by choosing $\langle (1,7),(1,2),(1,7)\rangle$ at time $40$ and $50$ or to cover two cycles by switching to $\langle (7,0),(5,2),(7,0)\rangle$ at $51$ and $59$ respectively. Both choices results in maximisation of the primary objective, i.e., the maximization of the number of cycles covered within $H=29$.
However, the second objective of $\textsf{DT*}$ prefers the solution which has shorter length for the last cycle covered, as explained earlier. Therefore, the solver decides to switch to $\langle (7,0),(5,2),(7,0)\rangle$ as it has shorter length for the last cycle covered.
This plan results in the completion of $7$ cycles till timestamp $85$, as shown in Figure~\ref{timeline2}(c).

The solution generated by $\textsf{DT*}$ was dependent on the choice of horizon $H$. If we take $H=49$, then there will be two choices at timestamp $4$ which will result in maximization of number of cycles within $H=49$. The choices are as follows:
(i) Choose the cycle $\langle (1,7),(1,2),(1,7)\rangle$ and cover $4$ cycles on timestamps $19,29,39,49$ respectively. (ii) Choose the cycle $\langle (7,0),(5,2),(7,0)\rangle$ to cover $4$ cycles at timestamps $27,35,35,51$. Again the solver prefers solution which minimizes the length of the last cycle. Therefore, generated solution will be to switch to $\langle (7,0),(5,2),(7,0)\rangle$. The plan generated is shown in Figure~\ref{timeline2}(d). This plan with $H=49$ will be equivalent to plan generated by $\greedyone$ algorithm in terms of the number of cycles completed within timestamp $85$.
\end{example}

The above example shows that a greedy algorithm may outperform $\textsf{DT*}$ in a long duration in some cases. However, our experimental results establish that such instances are rare, and overall $\textsf{DT*}$ offers superior performance than both the greedy algorithms.
}


\section{Evaluation}
\subsection{Experimental Setup}

\begin{figure}
\begin{minipage}{0.45\linewidth}
\centering
\includegraphics[height=3.7cm,scale=1]{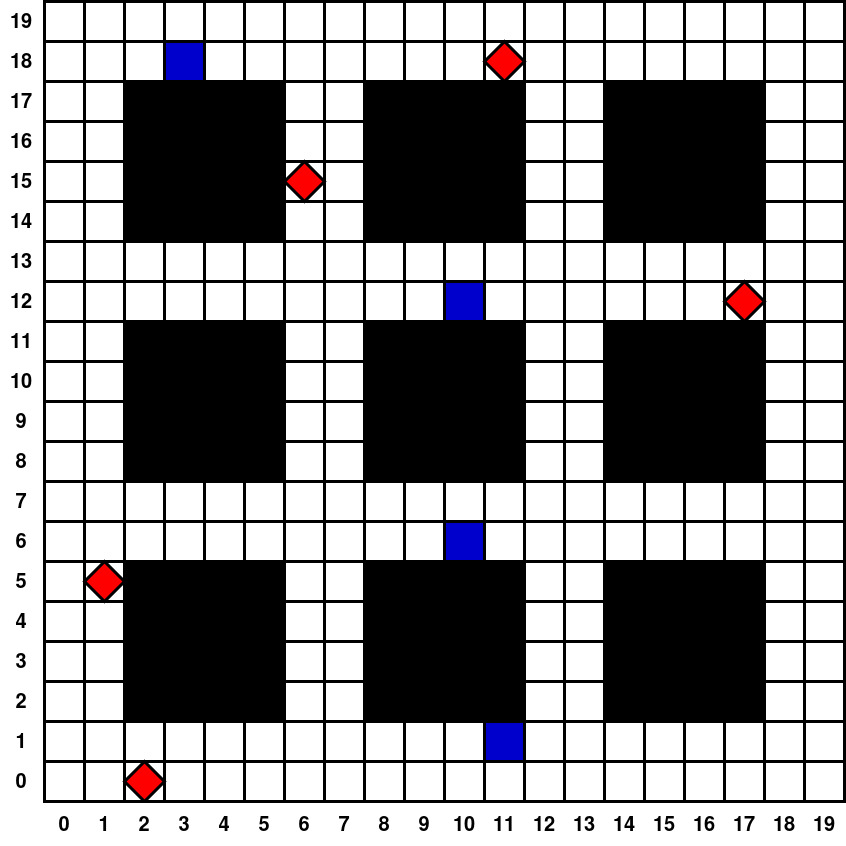}

\end{minipage}%
\qquad
\begin{minipage}{0.45\linewidth}
\centering
\includegraphics[width=0.9\linewidth]{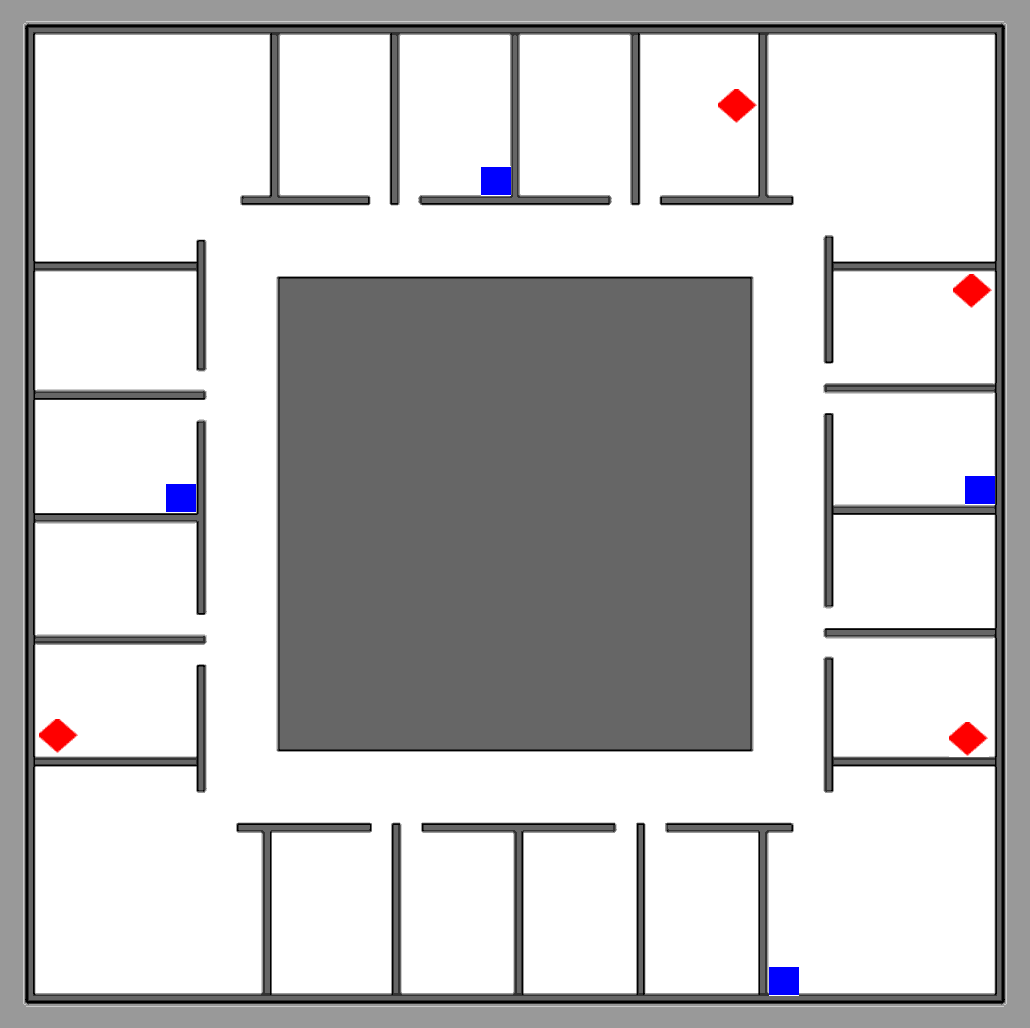}

\end{minipage}
\caption{(a) A $20\times20$ warehouse (b) A $100\times100$ Office\_h workspace with marked pickup and drop locations }
\label{fig7}
\end{figure} 
In this section, we present the experimental results on a pick and drop application in a $20\times20$ warehouse workspace as shown in Figure~\ref{fig7}(a) and a $100\times 100$ office workspace, taken from ~\cite{Rseg} as shown in Figure~\ref{fig7}(b).
To create different environments out of the warehouse workspace, we mark the locations in the grid, which could act as proposition locations. The environment descriptions for Figure~\ref{fig7}(a) are as follows:
$W_1$: Pickup locations: $(1,5)$, $(11,18)$, $(17,12)$ and Drop locations: $(3,18)$, $(10,6)$,$(10,12)$, $W_2$: $W_1$ + drop location $(11,1)$, 
$W_3$: $W_2$ + pickup location $(2,0)$ and $(6,15)$.

The LTL query that we use for the evaluation was introduced in Section~\ref{sec-example} and formally written as:
\resizebox{0.475\textwidth}{!}{$\phi \equiv \Box(\Diamond p\ \wedge \Diamond d)\ \wedge  \Box\ ((p \rightarrow X(\neg p \ \mathtt{U} \ d))\ \wedge (d \rightarrow X(\neg d\ \mathtt{U} \  p))$.}
In the above query, $p$ denotes a \emph{pickup
location} and $d$ denotes a \emph{drop location}.
By assuming some distribution over obstacle arrival rates, $A \sim \mathcal{N}(\mu,\,\sigma^{2})$, we change the workspace $W$ dynamically at various timestamps.

We implement our algorithm and the two greedy algorithms in C++. In the implementation, we use Z3~\cite{Z3} SMT solver as the back-end solver.
We carried out experiments also using Gurobi optimizer~\cite{gurobi} by modeling our problem as an appropriate Integer Linear Programming problem. However, in our experiments, Z3 consistently outperformed Gurobi in terms of computation time. Thus, we present our results using Z3 as the back-end solver only.
The results shown in this section have been obtained on a system with 3.2\,GHz octa-core processor with 32\,GB RAM.
We have repeated every experiment $50$ times to present the results. 


The operation of the robot is divided into path planning and path execution. In our experiments, we assume to use Turtlebot, a widely used mobile robot in academic research.
As given in~\cite{turtlebot_spec}, the popular robot Turtlebot 2 takes around $1\si{\second}$ to cover $0.65\si{\meter}$. Assuming the size of each cell of the grid to be $65\si{\centi\meter} \times 65\si{\centi\meter}$, we can say that each valid $act \in Act$, takes about $1\si{\second}$ to execute. 
We consider $\mathtt{left, right, up , down}$ motion primitives in the set $Act$.

We find the value of $time_{comp}$ experimentally. 
We find that it is safe to consider the value of  $time_{comp}$ to be $1\si{\second}$ and $2\si{\second}$ for the warehouse and office workspaces, respectively.
The time taken by Greedy algorithms can be ignored safely.


\subsection{Results}

We evaluate our planning framework by varying the following:
(i) the number of the suffix cycles in $G_r$,
(ii) the obstacle arrival rate,  
(iii) the size of the workspace and (iv) the obstacle density in the workspace.
For the evaluation, we vary one of the above-stated parameters while keeping the other parameters constant and empirically compare the performance of $\textsf{DT*}$ and that of the greedy algorithms.

The choice of horizon length $H$ is dependent on the obstacle arrival rate, $A \sim \mathcal{N}(\mu,\,\sigma^{2})$.
If $H$ is larger than the frequency with which the locations become unavailable, then the plan may get invalidated before completely getting traversed by the robot.
On the other hand, if $H$ is very small, then the solver may not consider some of the cycles just because those cycles could not be covered within the horizon. In our experiments, we kept the horizon length $H$ to be the mean $\mu$ of $A \sim \mathcal{N}(\mu,\,\sigma^{2})$.
\begin{figure}
    \centering
    \includegraphics[width=\linewidth]{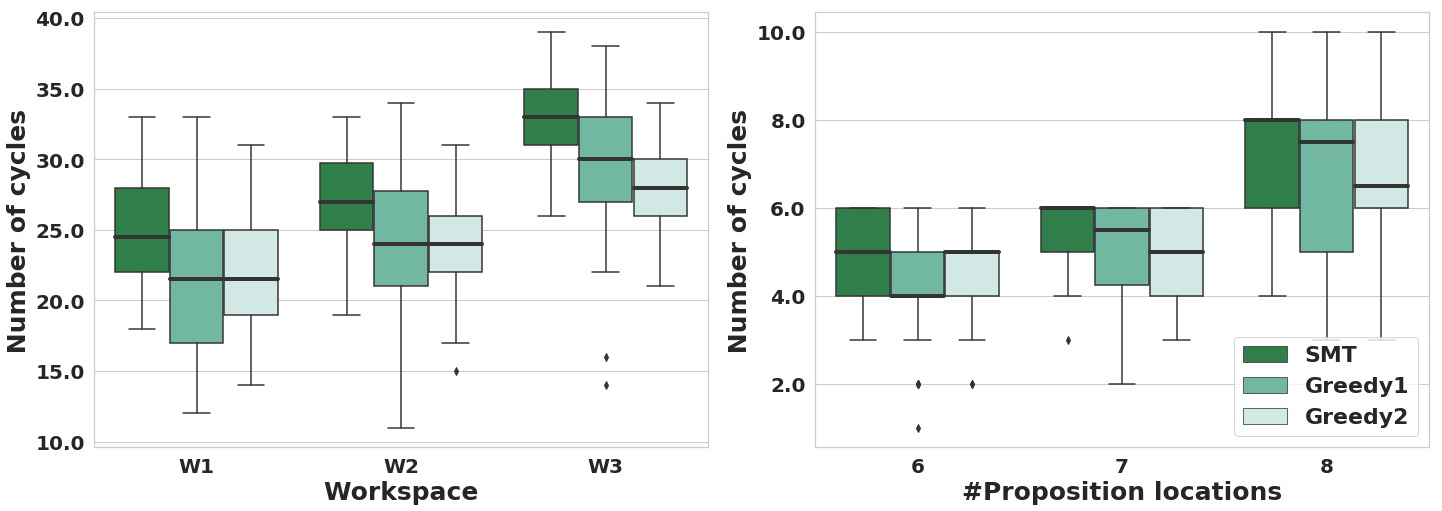}
    \caption{Performance of the algorithm with increasing proposition locations in (a) Warehouse and (b) Office\_h workspace} 
    \label{fig9}
\end{figure}
\subsubsection{Varying the number of proposition locations in a workspace}

We evaluate the performance of the algorithms by increasing the number of the proposition locations in the workspaces shown in Figure~\ref{fig7}(a) and Figure~\ref{fig7}(b) respectively. 
Figure~\ref{fig9} shows that the performance of all three algorithms improves with the increase in the number of proposition locations as the length of the suffix cycle decreases. However, in all the cases, $\textsf{DT*}$ outperforms the greedy algorithms.
There were a few cases where the plans generated by \greedyone algorithm were superior compared to the plans generated by $\textsf{DT*}$. 

For Figure~\ref{fig9}(a) the total planning time was $500\si{\second}$ timestamps and  obstacle arrival rate being $A \sim \mathcal{N}(100,20)$. And for Figure~\ref{fig9}(b), the total planning time was $1000\si{\second}$ and $A \sim \mathcal{N}(500,50)$.


\subsubsection{Varying obstacle parameters in the environment}


    
\begin{figure}
    \centering
    \includegraphics[width=\linewidth]{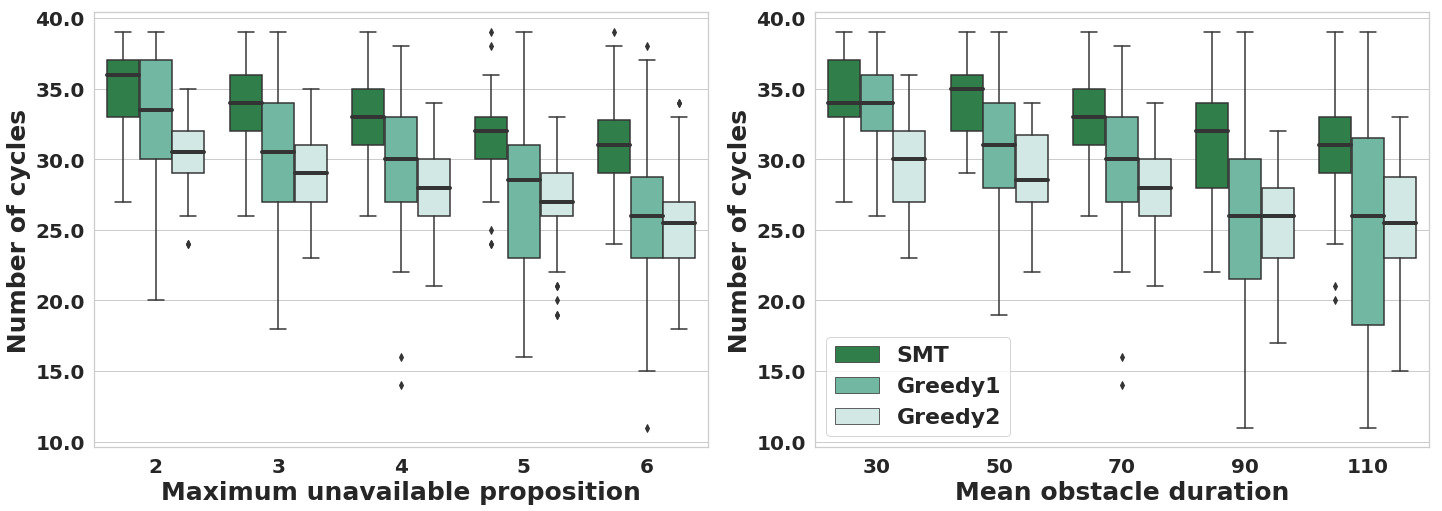}
    \caption{Performance of the algorithms (a) when we increase the maximum number of proposition locations unavailable at each environment change,
 (b) when we increase the duration of proposition locations' unavailability.}
    \label{fig10}
\end{figure}

One important parameter affecting the number of cycles traversed is the maximum number of proposition locations that become unavailable in each environment change. Figure~\ref{fig10}(a) shows that \greedyone and \greedytwo algorithm suffers significantly when the maximum number of proposition locations that become unavailable per environment change increases.  

Another important parameter is the duration of unavailability of the locations.
Here we have assumed the duration to follow a distribution $D \sim \mathcal{N}(\mu,\,\sigma^{2})$, the duration increases from $(30,10)$ to $(110,30)$. 
Figure~\ref{fig10}(b) shows that both greedy algorithms suffer when this obstacle duration increases. \greedyone algorithm suffers more when the unavailability duration of the proposition locations is high. The reason is that even if a shorter prefix path is available for longer cycles, it still prefers the cycle with shorter suffix and longer prefixes (incorporating high wait times). 

All of the above stated experiments are carried out on workspace $W_3$ with a total planning time of $500\si{\second}$.

\subsubsection{Varying grid size}

We scale our workspace from $20\times 20$ up to $50\times 50$ by keeping the number of proposition locations constant but increasing the cycle lengths proportionally by increasing the distance between the proposition locations. That is, we increase the cycle length(s) as grid size increases, which leads to fewer cycles covered within a fixed duration by all the algorithms. 
Figure~\ref{fig11}(a) shows the performance of the three algorithms. As expected, $\textsf{DT*}$ consistently outperforms the greedy algorithms. The results show that
\greedytwo algorithm specifically performs poorly on larger workspaces $40\times40$ and $50\times50$. This happens due to the fact that the prefix path(s) increases with the scaling of the grid and \greedytwo starts preferring shorter prefixes with longer corresponding suffix cycles. That is \greedytwo algorithm starts generating closer sub-optimal solutions.





\subsubsection{Effect of objective functions}
\begin{figure}
    \centering
    \includegraphics[width=\linewidth]{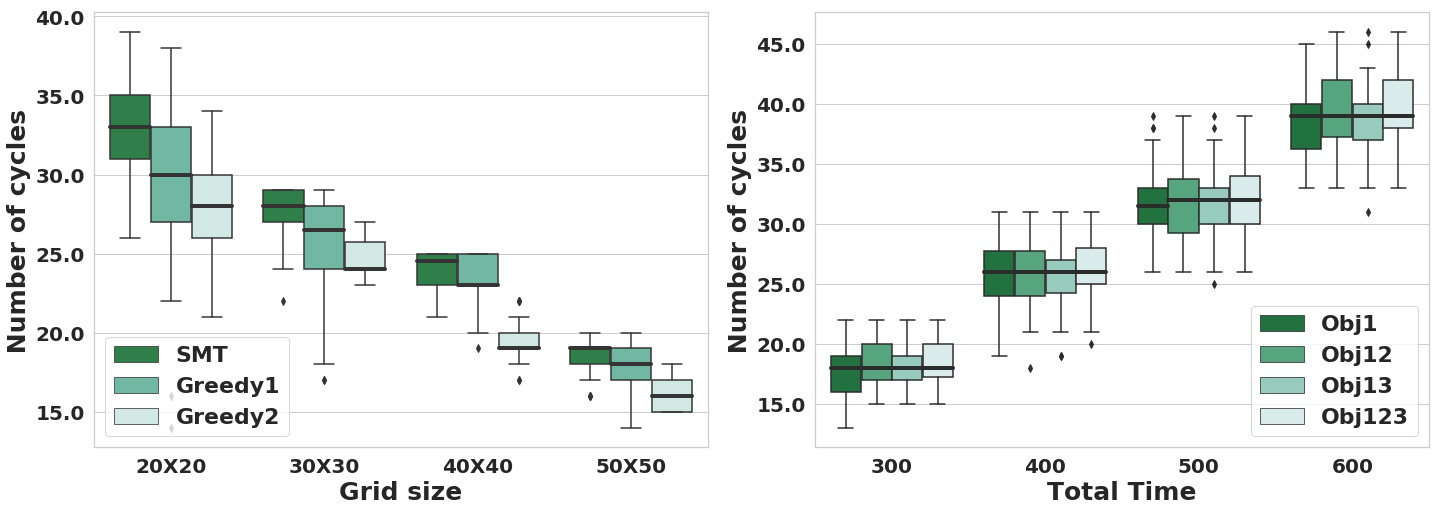}
    \caption{(a) Performance of algorithm for increasing size of the workspaces with total planning time=$500\si{\second}$. (b) Results for various combinations of the objective functions for different total planning times.}
    \label{fig11}
\end{figure}

 In the same dynamic environment, we generate decision sequences by $\textsf{DT*}$ by giving it some combination of objective functions 1, 2, and 3. Figure~\ref{fig11}(b) shows that the usage of all three objectives yields the best results. In all the combinations, the primary objective was always to maximize the number of cycles. The mentioned experiment is carried out on workspace $W_3$ by increasing the total planning time.

\subsubsection{Computation Time}
Table~\ref{tab1} shows the effect of changing the number of proposition locations and the horizon length on the overall time taken by $\textsf{DT*}$.
Table~\ref{tab1} also shows that as the initial length of the smallest cycle in the workspace decreases, the solver takes more time to solve the constraints, as more cycles could be completed within the same horizon. 

Through these experiments, we demonstrate that though $\textsf{DT*}$ involves solving an optimization problem, its computation time is not significant. Overall, our $\textsf{DT*}$ algorithm outperforms the greedy algorithms in terms of the total number of cycles covered within a given duration in the majority of cases.

\begin{table}[t]
\caption{Computation time of $\textsf{DT*}$}
\label{tab1}
\resizebox{\linewidth}{!}{%
\begin{tabular}{|l|c|c|c|c|c|}
\hline
Workspace & \# Propo & Horizon & \multicolumn{3}{c|}{Computation time (\si{\second})}\\
\cline{4-6}
 & sition & length &$\mathtt{dy\_cost}$ & $\mathtt{plan\_in\_H}$ & Total \\
\hline
\textbf{$W_1$} & \textbf{6}& 100& 0.019 $\pm$ 0.004& 0.046 $\pm$ 0.016 & 0.063 $\pm$ 0.016\\
\hline
\textbf{$W_2$} & \textbf{7}& 100& 0.023 $\pm$ 0.005 & 0.103 $\pm$ 0.03 & 0.125 $\pm$ 0.032\\
\hline
\textbf{$W_3$} & \textbf{9}& 100& 0.05 $\pm$ 0.010& 0.224 $\pm$ 0.051 & 0.268 $\pm$ 0.054\\
\hline
\hline
$W_3$ & 9& \textbf{50} & 0.035 $\pm$ 0.007 & 0.016 $\pm$ 0.011 & 0.045 $\pm$ 0.009 \\
\hline
$W_3$ & 9&\textbf{70} & 0.035 $\pm$ 0.06 & 0.06 $\pm$ 0.024 & 0.099 $\pm$ 0.034 \\
\hline
$W_3$ & 9&\textbf{100} & 0.05 $\pm$ 0.010& 0.224 $\pm$ 0.051 & 0.268 $\pm$ 0.054 \\
\hline
$W_3$ & 9&\textbf{120} & 0.039 $\pm$ 0.01 & 0.396 $\pm$ 0.115 & 0.421 $\pm$ 0.092\\
\hline
\hline
Office\_h & \textbf{6} &500 & 0.499 $\pm$ 0.125 & 0.010 $\pm$ 0.004 & 0.509 $\pm$ 0.128 \\
\hline
Office\_h &\textbf{7} &500 & 0.659 $\pm$ 0.151 & 0.045 $\pm$ 0.023 & 0.702 $\pm$ 0.163 \\
\hline
Office\_h &\textbf{8} &500& 0.863 $\pm$ 0.239 & 0.189 $\pm$ 0.119 & 1.04 $\pm$ 0.288\\
\hline
\end{tabular}
}

\resizebox{\linewidth}{!}{%
\begin{tabular}{|l|c|c|c|c|c|}
\hline
\hline
Workspace & Smallest & Horizon & \multicolumn{3}{c|}{Computation time (\si{\second})}\\
\cline{4-6}
 & Cycle & length &$\mathtt{dy\_cost}$ & $\mathtt{plan\_in\_H}$ & Total \\
\hline
$W_3$ & \textbf{8}& 100& 0.049 $\pm$ 0.01& 0.257 $\pm$ 0.071 & 0.296 $\pm$ 0.074\\
\hline
$W_3$ & \textbf{12}& 100& 0.05 $\pm$ 0.01& 0.224 $\pm$ 0.051 & 0.268 $\pm$ 0.054\\
\hline
$W_3$ & \textbf{16}& 100& 0.052 $\pm$ 0.015& 0.196 $\pm$ 0.065 & 0.239 $\pm$ 0.071\\
\hline
\end{tabular}
}
\end{table}

\subsection{ROS+Gazebo Experiment}
We provide a Gazebo simulation of the decision sequence generated by $\textsf{DT*}$ for the example explained in~\ref{sec-example} section, as supplementary material. The timeline in Figure~\ref{fig2}(c) captures the plan that was generated by $\textsf{DT*}$. The proposition locations $P1$ and $P3$ become unavailable at timestamp $10$ when they are blocked by other agents. These agents communicate the duration for which the proposition locations will be blocked to the robot.
The robot uses an SMT solver, which takes $1\si{\second}$  to generate a plan at timestamp $11$. The robot then traverses this generated plan to complete $4$ cycles up to timestamp $50$. In our Gazebo simulation we use Turtlebot~\cite{turtlebot_spec} as the robot and AMCL localization method~\cite{ThrunFBD01} provided by Rviz.


\section{Related Work}
\label{sec-related}
LTL is a popular logical language for capturing complex requirements for robotic systems. Several researchers have addressed the motion planning problem from LTL specification in the past. The techniques to solve the problem includes graph-based techniques~\cite{UlusoySDBR13}, sampling-based technique~\cite{KaramanF09,BhatiaKV10}, Constraint solving based techniques~\cite{WolffTM14} and classical planning extend with the capability to deal with temporal logic specifications~\cite{PatriziLGG11}. 
For a detailed review of the LTL motion planning literature, the readers are referred to the survey paper by Plaku and Karaman~\cite{PlakuK15}.

Planning in a dynamic environment for rechability specification (reaching a goal location avoiding dynamic obstacles) has been widely studied in graph based settings \cite{dy1, dy2,dy3,dy4,dy5}.
For temporal logic specification, reactive synthesis for GR(1) subset of LTL has been undertaken in~\cite{Kress-GazitFP09,WongpiromsarnTM12}. In this approach, a reactive controller is synthesized to enable the robot to react to the inputs coming from the environment. The reactive synthesis is performed based on some assumptions on the workspace. How to deal with the situations when the assumptions on the workspace get violated has been addressed in ~\cite{LivingstonPJM13,WongEK14}.  Though our work can be seen as a reactive synthesis problem in a dynamic workspace, we do not take the route of reactive synthesis as it generally suffers from the lack of scalability.

Our algorithm is based on a reduction of the problem to an SMT solving problem. The SMT-based approach has been adopted in solving various  motion planning problems, for example, motion planning for reachability specification~\cite{HungSTLZWG14,SahaRKPS16}, motion planning for LTL specification~\cite{SahaRKPS14,ShoukryNSSSPT16,ShoukryNBSSSPT17}, energy-aware temporal logic motion planning~\cite{KunduS19}, integrated task and motion planning~\cite{NedunuriPMCK14,WangDCK16}, centralized and decentralized motion planning for mobile robots ~\cite{GavranMS17,DesaiSYQS17}, and multi-robot coverage planning~\cite{DasS18}.
We, for the first time, apply an SMT-based technique to address the online LTL motion planning problem in a dynamic environment.
\newcommand{\Mypm}{\mathbin{\tikz [x=1.4ex,y=1.4ex,line width=.1ex] \draw (0.0,0) -- (1.0,0) (0.5,0.08) -- (0.5,0.92) (0.0,0.5) -- (1.0,0.5);}}%
\noindent\setlength\tabcolsep{3pt}%

\section{Conclusion}
\label{sec-conclusion}
In this paper, we have presented $\textsf{DT*}$, an SMT-based approach to solving the LTL motion planning problem in a dynamic environment, which uses $\textsf{T*}$~\cite{KhalidiSaha18} as the backbone. 
We have shown empirically that $\textsf{DT*}$ is capable of generating a superior execution plan in terms of the maximization of task completion. 
Our approach incurs computation time which is required to solve constraint satisfaction problems online by an SMT solver. Despite this computational disadvantage, the average gain in the number of loops completed within a fixed duration is significant compared to suitably crafted greedy algorithms. Our future work includes extending this work for multi-robot systems and computing the ideal horizon length $H$ by learning the distribution over the obstacle arrival rate.


\bibliographystyle{IEEETran}
\bibliography{ref,bibli,references}

\end{document}